\documentclass[10pt,twocolumn,letterpaper]{article}
\usepackage[pagenumbers]{cvpr}
\usepackage{amsthm,bm,mathtools,tabularx,array,placeins}
\usepackage[T1]{fontenc}
\usepackage{microtype}
\newtheorem{theorem}{Theorem}
\newtheorem{proposition}{Proposition}

\newtheorem{corollary}{Corollary}
\theoremstyle{remark}
\newcommand{\E}{\mathbb E}
\newcommand{\PP}{\mathbb P}

\newcommand{\KL}{\operatorname{KL}}

\newcommand{\Cov}{\operatorname{Cov}}
\newcommand{\tr}{\operatorname{tr}}
\newcommand{\diag}{\operatorname{diag}}

\newcommand{\dd}{\mathrm d}

\definecolor{cvprblue}{rgb}{0.21,0.49,0.74}
\usepackage[pagebackref,breaklinks,colorlinks,allcolors=cvprblue]{hyperref}
\def\confName{CVPR}
\def\confYear{2027}
\title{Growing Apart on the Same Data: \\Training-Time Double Descent of Replica Discrepancy}
\author{
Yifan Luo \quad Bin Dong\\
Peking University\\
{\tt\small luoyf@pku.edu.cn \quad dongbin@math.pku.edu.cn}
}
\begin{document}
\maketitle
\begin{abstract}
When two neural networks learn from the same data, we expect their predictions to grow more alike. However, we identify training regimes in which continued training temporarily drives their predictions further apart. In experiments with ResNet-18 and ViT-Tiny on CIFAR-10 and ImageNet-100, we observe a decrease--increase--decrease trajectory in symmetric KL between independently initialized replicas, with all remaining training randomness shared. We analyze two-layer linear networks with both layers trained by gradient flow on inputs with strong and weak directions. In the limit of a vanishing weak-input scale, we prove that replicas follow a common learning curve with initialization-dependent time shifts, yielding early agreement, intermediate separation, and late agreement. This characterization predicts relative learning times from the initial weights. The same phase mechanism also permits improving or worsening population cross-entropy, depending on what the replicas learn in common. Our findings show that early prediction agreement can mask the influence of initialization, which may re-emerge and become amplified during intermediate stages of training.
\end{abstract}

\section{Introduction}
Training on the same examples should bring two neural networks toward the same predictions. However, previous work in image classification has shown that networks trained on the same data can learn different decision boundaries and make different predictions \cite{somepalli2022,bhojanapalli2021,summers2021}. In this work, we observe \emph{training-time double descent of replica discrepancy} even when initialization is the only difference between runs.

\begin{figure}[t]\centering
\includegraphics[width=\linewidth]{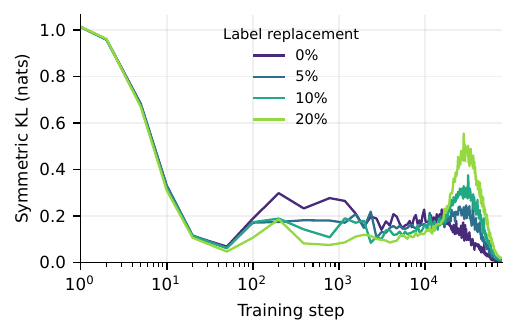}
\caption{\textbf{Early agreement can be followed by renewed separation.} Symmetric KL between ResNet-18 replicas trained on CIFAR-10 with different label replacement ratios.}\label{fig:mainphenomenon}
\end{figure}

In our experiments, we study pairs of independently initialized networks. They share the training data, observed labels, minibatch order, augmented inputs, and all remaining training randomness. We measure the symmetric KL divergence between the replicas throughout training. We observe this double-descent pattern in ResNet-18 and ViT-Tiny experiments on CIFAR-10 and ImageNet-100 under different label conditions and training recipes. \Cref{fig:mainphenomenon} illustrates the pattern for ResNet-18 on CIFAR-10. Discrepancy drops sharply during early training, rises after this initial alignment, and falls again at later stages.

To identify what contributes to the resurgence, we track the prediction discrepancy between replicas on each sample throughout training. Averaging these per-sample discrepancies recovers the model-level trajectory. We decompose changes in this average into positive and negative per-sample contributions. The aggregate discrepancy rises when increases outweigh decreases in magnitude, and falls when this balance reverses. Local response calculations then show how shared updates can generate positive contributions even where the replicas currently agree. 

Guided by these analyses and prior work on staged learning \cite{saxe2014,pesme2023,atanasov2022}, we analyze a fully trained two-layer linear network with strong and weak input directions. Strong-direction fitting first aligns predictions. Learning then amplifies the weak response, exposing initialization-dependent timing differences. For gradient flow with a vanishing weak-input scale, we prove that replicas follow a common learning curve with phases determined by their raw initial weights. These phases predict the relative timing of weak-direction learning, producing early agreement, intermediate separation, and late agreement. Timing measurements and function-matched controls test the predicted learning phases and response amplification. This phase mechanism also clarifies how predictive reproducibility relates to prediction quality. Replicas can exhibit the same discrepancy pattern while population risk either improves or worsens, depending on what they learn in common.

Our contributions are threefold.
\begin{itemize}
\item \textbf{A counterintuitive phenomenon.} We identify and report training-time double descent of replica discrepancy. Even when runs differ only in initialization, continued training can drive their predictions apart after an early period of agreement.
\item \textbf{Mechanism analysis.} We decompose the aggregate discrepancy dynamics into per-sample prediction changes and characterize how shared updates can regenerate discrepancy.
\item \textbf{Theory and validation.} We derive an initialization-dependent phase limit for fully trained two-layer linear networks and test the mechanism through timing predictions and function-matched controls.
\end{itemize}

These findings give a dynamical view of predictive reproducibility across initializations, complementing studies of sensitivity to training randomness \cite{damour2022,butterfly2025,ly2026}. Early prediction agreement can coexist with different future learning dynamics, so reproducibility can weaken and recover over a training run. Our phase analysis connects this evolution to changing prediction responses and the relative timing of learning, explaining how a common training procedure can suppress and later amplify initialization-dependent variation.

\section{Related work}\label{sec:related}
\paragraph{Staged learning and initialization.}
Staged mode learning and initialization-dependent delays arise in deep-linear and diagonal networks \cite{saxe2014,gissin2020,jacot2022,pesme2023}, with recent analyses treating deep ReLU saddle escape and staged dynamics in convolutional and attention architectures \cite{bantzis2026,zhang2026}. Even et al.\ \cite{even2023} characterize how stochasticity and step size alter diagonal-network dynamics. In small-initialization models, silent alignment separates changes in learned directions from later growth in response scale \cite{atanasov2022,martin2026}. Conserved layer imbalance and initialization determine feature-learning regimes \cite{du2018,azulay2021,kunin2024,xu2025}, while scaling can enforce lazy dynamics \cite{chizat2019}. Related temporal structure appears in simple-pattern learning before memorization \cite{arpit2017,han2025} and in delayed generalization through grokking \cite{power2022,kumar2024}. We derive relative transition times from raw initialization to explain how shared learning stages produce differences between replicas.

\paragraph{Double descent.}
Double descent describes nonmonotonic risk as model complexity or training time increases \cite{belkin2019,nakkiran2020}. Epoch-wise analyses explain risk resurgence through superposed bias--variance tradeoffs and features learned at different rates \cite{heckel2021,stephenson2021,pezeshki2022}. Other decompositions isolate data, label, and initialization variance \cite{adlam2020,dascoli2020}, while random-feature and linear-model analyses characterize time-dependent generalization under SGD \cite{liu2022,atanasov2025,bordelon2026}. Our object is the relative evolution of two predictors throughout a common training procedure. Initialization-dependent phases produce separation and renewed alignment even in a regime where classification error stays zero during weak learning (\cref{sec:twolayer}).

\paragraph{Predictive reproducibility.}
Image classifiers can differ across retrainings despite similar accuracy \cite{bhojanapalli2021,summers2021}, and decision-boundary reproducibility varies with width \cite{somepalli2022}. Churn-reduction methods encourage consistency across runs or model updates \cite{bahri2021,adam2023}. Underspecification connects such variability to deployment behavior \cite{damour2022}. Classification disagreement also relates to test error under calibration conditions \cite{jiang2022,kirsch2022}, with empirical and theoretical extensions to distribution shift \cite{baek2022,lee2023,saxena2024}. With independent training samples, agreement bounds characterize aggregation and architecture growth \cite{eaton2026}.

Training-time studies reveal early stabilization to SGD noise \cite{frankle2020,fort2020}. Alt{\i}nta\c{s} et al.\ \cite{butterfly2025} vary the time of a controlled perturbation and measure its effect on the resulting trajectories and endpoints. Ly and Gong \cite{ly2026} study limits to endpoint reproducibility through basin geometry. We follow pairs with different initial weights and shared subsequent randomness throughout training. Their early prediction agreement can be followed by intermediate separation and later renewed agreement, linking reproducibility to the relative timing of learning.

\section{Double descent of replica discrepancy}\label{sec:setting}
We first define a controlled comparison in which training runs differ only in initialization and specify how their prediction discrepancy is measured. We then present double-descent trajectories for CNNs and vision transformers on CIFAR-10 and ImageNet-100 across label sources and training recipes.

\begin{figure*}[t]\centering
\includegraphics[width=\linewidth]{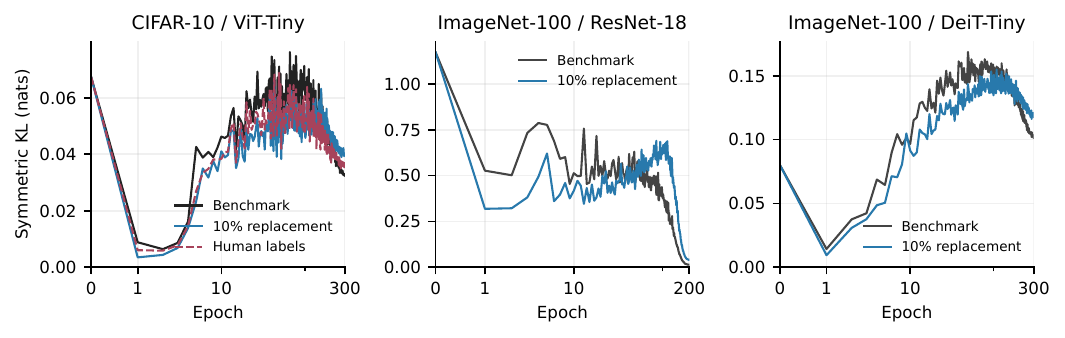}
\caption{\textbf{The resurgence extends beyond the primary CNN recipe.} Unsmoothed integer-epoch, full-training-set KL for strongly regularized ViT-Tiny/4 on CIFAR-10, ImageNet-100 ResNet-18, and ImageNet-100 DeiT-Tiny/16. The epoch axes are linear through epoch 1 and logarithmic thereafter. All runs use five-epoch warmup and finish their full budgets. Each condition has one coupled initialization pair.}\label{fig:mainvision}
\end{figure*}

\subsection{Controlled replicas and measurements}
We train two replicas that differ only in initialization, sharing a fixed training set $\mathcal D=\{(x_i,\widetilde y_i)\}_{i=1}^N$, architecture, optimizer, minibatches, and transformed input tensors. Forward randomness, including stochastic depth and mixed targets where used, is also shared.

For probabilities $p_{jt}(x)=\operatorname{softmax}z_{jt}(x)$, define
\begin{equation}
 S_Q(t)=\frac12\E_{X\sim Q}\!\left[\KL(p_{0t}\Vert p_{1t})+\KL(p_{1t}\Vert p_{0t})\right].
 \label{eq:mainmetric}
\end{equation}
Unless stated otherwise, $Q$ is uniform over all fixed, unaugmented training evaluation views. Held-out discrepancy uses a separate input set. Each evaluation includes every example in the selected set. For softmax outputs, symmetric KL also admits the exact expression $S(p_0,p_1)=\tfrac12(p_0-p_1)^T(z_0-z_1)$.

\subsection{CIFAR-10 experiments with ResNet-18}
We train CIFAR-adapted ResNet-18 networks \cite{he2016} with GroupNorm \cite{wu2018} from scratch, using benchmark labels and 5\%, 10\%, or 20\% label replacement. All image experiments use five epochs of warmup followed by cosine decay. Training details and initialization/normalization ablations are in \cref{sec:primarysupp,sec:seedablation,sec:bnablation}.

\Cref{fig:mainphenomenon} shows training-time double descent, with an initial decline, an intermediate resurgence, and a late decline in replica discrepancy. For ResNet-18 on CIFAR-10 with 10\% label replacement, discrepancy drops sharply over the first six epochs, rises between epochs 6 and 81, and declines again from epoch 81 through epoch 200. This behavior is counterintuitive. Continued training on identical minibatches drives the replicas apart after initially bringing their predictions closer, even though only initialization differs.

\subsection{Ablation across architectures and data}
We next vary the architecture, dataset, and label source to examine their influence on the double-descent trajectory.

\Cref{fig:mainvision} presents ViT-Tiny/4 on CIFAR-10 \cite{krizhevsky2009} and both ResNet-18 and DeiT-Tiny/16 on ImageNet-100 \cite{deng2009,tian2020}. ImageNet-100 contains 126,689 training images and 5,000 validation images over a fixed 100-class subset. All models train from scratch. Detailed architecture and training configurations appear in \cref{sec:definitions,sec:imagenetprotocol}.

On CIFAR-10, ViT-Tiny/4 shows a resurgence with 10\% synthetic label replacement, CIFAR-10N human annotations \cite{wei2022}, and benchmark labels. On ImageNet-100, both ResNet-18 and DeiT-Tiny/16 show an initial decline, an intermediate resurgence, and a later decline under both benchmark labels and 10\% label replacement. Transformer replicas retain measurable discrepancy at the end of training. Additional label-source and training-recipe comparisons appear in \cref{sec:vit,sec:imagenet}.

In the CIFAR-10 ResNet-18 experiments, added label noise makes the resurgence more pronounced. The intermediate peak rises as replacement increases from 0\% to 20\%, with the highest peak at 20\% (\cref{fig:mainphenomenon}). For DeiT-Tiny/16 on ImageNet-100, 10\% replacement instead lowers and delays the intermediate peak relative to benchmark labels (right panel of \cref{fig:mainvision}). ViT-Tiny/4 also exhibits a clear double-descent trajectory on CIFAR-10 with benchmark labels and no injected label noise.

\section{Interpreting the phenomenon}\label{sec:analysis}
We decompose the model-level discrepancy trajectory into per-sample changes to identify what contributes to the resurgence. We examine sample-fitting times and forgetting, then study how models that currently agree can respond differently to shared minibatches. All diagnostics in this section use the same GroupNorm ResNet-18 experiment on CIFAR-10 with 10\% label replacement. These observations guide the subsequent theoretical analysis.

\begin{figure}[t]\centering
\includegraphics[width=\linewidth]{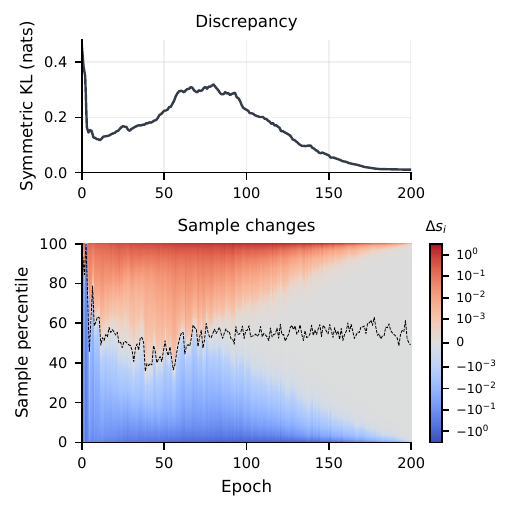}
\caption{\textbf{Connecting sample changes to model discrepancy.} All 50,000 training images from the CIFAR-10 ResNet-18 experiment with 10\% label replacement. Each sample's KL is smoothed over five centered integer-epoch observations, using available observations at the edges. The upper plot shows mean discrepancy. The lower plot shows adjacent sample changes, sorted independently into percentiles at each interval midpoint. Red indicates increases and blue indicates decreases. The thin dashed line marks the nonpositive fraction.}\label{fig:mainsamples}
\end{figure}

\subsection{What the aggregate rise measures}
Let $s_i(t)$ denote the symmetric KL between the replicas on training image $i$ at observation $t$. The model-level discrepancy is their mean, $S(t)=N^{-1}\sum_i s_i(t)$. Changes between any two observations therefore satisfy
\begin{equation}
 \begin{gathered}
 \Delta S=U-V,\\
 U=\frac1N\sum_i[\Delta s_i]_+,\quad
 V=\frac1N\sum_i[-\Delta s_i]_+.
 \end{gathered}
 \label{eq:mainsample}
\end{equation}
We apply this decomposition to all 50,000 training images in the CIFAR-10 ResNet-18 experiment with 10\% label replacement. Positive and negative sample changes contribute $U$ and $V$ to the aggregate trajectory. Their balance determines how the combined prediction changes increase or decrease the model-level discrepancy.

\Cref{fig:mainsamples} reveals how the balance of sample-level changes follows the model-level trajectory. After early alignment, positive changes become more prominent as discrepancy rises. Negative changes regain the larger aggregate contribution during the later decline, and both weaken toward the end of training. The positive share of total change magnitude, $U/(U+V)$, exceeds $1/2$ precisely when the mean KL increases and falls below $1/2$ when it decreases. The resurgence thus reflects a shift toward positive contributions among samples whose discrepancies continue to move in both directions.

\subsection{Learning times and forgetting}
\begin{figure}[t]\centering
\includegraphics[width=\linewidth]{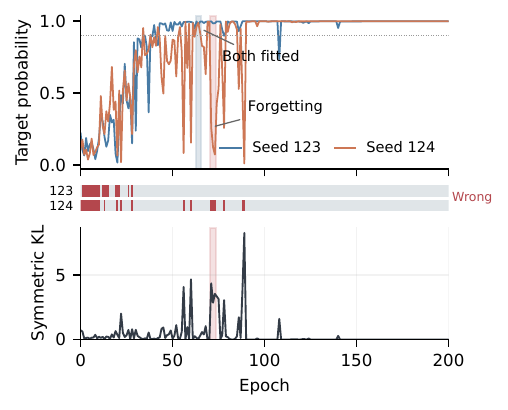}
\caption{\textbf{Asymmetric forgetting after joint fitting.} A correctly labeled CIFAR-10 image from ResNet-18 trained with 10\% label replacement. The upper plot shows target-class probabilities, with red strips marking incorrect predictions. The lower plot shows symmetric KL for the same image.}\label{fig:sampleforgetting}
\end{figure}

Example forgetting occurs when a previously correct prediction becomes incorrect \cite{toneva2019}. Replicas can jointly fit a sample and later diverge through asymmetric forgetting. \Cref{fig:sampleforgetting} shows a correctly labeled CIFAR-10 image from the ResNet-18 experiment with 10\% label replacement. Both replicas predict it with high confidence at epochs 63--65. Seed 124 subsequently undergoes forgetting at epochs 71--73 while seed 123 maintains the correct prediction, and their KL rises. Their predictions align again later in training.

The full-sample analysis of the same GroupNorm ResNet-18 experiment with 10\% label replacement considers correctly labeled images with one correct replica at the discrepancy peak. Of this group's KL, 85.12\% comes from images that the currently incorrect replica had previously predicted with observed-label probability at least 0.9 for three consecutive epochs. For 50.73\% of the same group's KL, the other replica remains correct after both had simultaneously sustained this high-confidence condition. Initialization-dependent differences therefore become visible in how previously fitted samples are retained or forgotten (\cref{sec:matchedprimary}).

\subsection{Shared updates and renewed discrepancy}
We examine how shared updates can regenerate discrepancy after prediction alignment. We derive their local effect on KL, then test the predicted response using function-equivalent ResNet parameterizations.

\paragraph{Shared updates can separate equal predictions.}
To isolate this effect, consider two models that currently agree on fixed query inputs, with common probability $p(x)$. Let $J_j(x)$ be the logit Jacobian, $g_{ji}$ a per-example CE gradient, $F_x=\diag(p)-pp^T$, and $h_i(x)=J_0(x)g_{0i}-J_1(x)g_{1i}$. For smooth local logits and a shared iid batch of size $B$, ordinary SGD gives
{\small
\begin{equation}
 \E_{\mathcal B}S_Q^+=\tfrac{\eta^2}{2}\E_{x\sim Q}\!\big[\bar h_x^TF_x\bar h_x+B^{-1}\tr(F_x\Sigma_x)\big]+O(\eta^3),
 \label{eq:mainlocal}
\end{equation}
}
Here $\bar h_x=\E_i h_i(x)$ and $\Sigma_x=\Cov_i h_i(x)$, with $F_x$ providing the local Fisher geometry of predictive KL \cite{martens2020}. Proposition~\ref{prop:local} in Appendix~\ref{sec:cnn} gives assumptions and proof. For ReLU networks, the expansion uses a locally fixed activation pattern. Equal functions can have different Jacobians and hence different update responses. The mean term can survive full-batch training, and the covariance contribution scales as $1/B$. Discrepancy regeneration requires the different Jacobian responses to the residual to remain visible through $F_x$.

To identify the source of the response difference, consider replicas that also agree on the training inputs. Their common cross-entropy residual is $r_i=p(x_i)-e_{\widetilde y_i}$. The response difference becomes
\begin{equation}
 \begin{aligned}
 h_i(x)&=[K_0(x,x_i)-K_1(x,x_i)]r_i,\\
 K_j(x,x_i)&=J_j(x)J_j(x_i)^T.
 \end{aligned}\label{eq:mainresponsekernel}
\end{equation}
Each $K_j$ is a logit neural tangent kernel \cite{jacot2018}, mapping a training residual to a first-order logit change on the query image. Equal current probabilities can thus coexist with different future prediction responses. The two terms in \cref{eq:mainlocal} measure the Fisher-weighted effects of the mean response and its batch-to-batch variability. This provides a local source of positive per-sample KL changes after alignment.

To test whether update geometry generates KL from identical predictions, we construct function-equivalent parameterizations from the same GroupNorm ResNet-18 experiment on CIFAR-10 with 10\% label replacement. A positive rescaling of hidden features and compensating classifier rescaling preserves predictions, while changing the parameter response to ordinary SGD \cite{dinh2017}. At epochs 20 and 200, the local quadratic prediction matches the mean KL over eight shared batches at step size $10^{-4}$ with relative errors $0.0054\%$ and $0.0021\%$, respectively. Queries and evaluation state are fixed, and updates are computed in FP64. The unchanged-parameterization control gives zero discrepancy. Its full configuration, protocol, and finite-step checks are in \cref{sec:matchedresponse}.

These controls show that matching predictions can coexist with different responses to shared training examples. The same minibatch can then regenerate discrepancy through the models' different parameterizations.

\section{Initialization-dependent learning phases}\label{sec:twolayer}
To explain how renewed separation develops over training, we analyze the joint evolution of predictions and their update responses in a fully trained two-layer linear network. Strong-direction fitting first aligns predictions, while amplification of a learned weak response exposes initialization-dependent learning times. In a vanishing weak-input-scale limit, we derive these times from the raw initial weights and characterize the resulting discrepancy trajectory. We also examine how the predictive value of the learned direction determines population risk.

\subsection{A fully trained two-layer model}
Consider $f_\theta(x)=u(v_1x_1+v_2x_2)$, with all three weights independently initialized as $\mathcal N(0,1)$ and trained. Inputs are $x=b(1,at)$, where $b,t\in\{-1,1\}$ are balanced signs and $a>0$ is a weak-input scale. Normalize logits by the reference sign $b$ and write $s=uv_1$, $m=auv_2$. For binary observed labels $\widetilde y_i\in\{-1,1\}$, let $T_i=(1+b_i\widetilde y_i)/2$ have empirical group means $q_\pm=\E_i[T_i\mid t_i=\pm1]=\bar q\pm c$. We condition on the fixed empirical design
\begin{equation}
 \bar q=1-\rho,\qquad 0<c<\rho<1/2,
 \label{eq:mainassumptions}
\end{equation}
giving logistic loss
\begin{equation}
 \mathcal L(s,m)=\tfrac12\!\sum_{r\in\{-1,1\}}
 [\log(1+e^{s+rm})-q_r(s+rm)].
 \label{eq:mainloss}
\end{equation}
For example, a fixed input-independent label-flip sample with true label $b$ can realize these conditional frequencies through finite-sample imbalance. The theorem conditions on the realized finite-sample correlation $c$ and keeps it fixed as $a\to0$, even when the corruption-generating population has zero correlation. A second instantiation below gives the weak direction genuine task information.

Deep-linear learning dynamics depend on the factorization and initialization \cite{saxe2014,kunin2024,xu2025}. Here both factors learn, inducing nonlinear dynamics in the coefficients. With $A=-\partial_s\mathcal L$ and $B=-\partial_m\mathcal L$, gradient flow gives
\begin{equation}
 \binom{\dot s}{\dot m}=
 \underbrace{\begin{pmatrix}u^2+v_1^2&av_1v_2\\av_1v_2&a^2(u^2+v_2^2)\end{pmatrix}}_{K}
 \binom{A}{B}.
 \label{eq:mainkernel}
\end{equation}
The shared output weight couples the directions. Eliminating the instantaneous strong response yields the weak Schur gain
\begin{equation}
 g=K_{mm}-K_{sm}^2/K_{ss}
   =a^2u^2\left(1+\frac{v_2^2}{u^2+v_1^2}\right).
 \label{eq:maingain}
\end{equation}
Strong fitting first brings $s$ near $s_0=\operatorname{logit}(\bar q)$ while $m$ is small. Later, $s$ tracks $s^*(m)$ defined by $\sigma(s^*+m)+\sigma(s^*-m)=2\bar q$. Let
\begin{equation}
 H(m)=\tfrac12[\sigma(s^*(m)+m)-\sigma(s^*(m)-m)].
 \label{eq:mainH}
\end{equation}
$H$ increases from zero to $\rho$, and the remaining weak residual is $c-H(m)$.

The layer-imbalance invariant \cite{du2018}, $D=u^2-v_1^2-v_2^2$, implies
\begin{equation}
 u^2=\tfrac12\left[D+\sqrt{D^2+4s^2+4m^2/a^2}\right].
 \label{eq:mainbalance}
\end{equation}
For bounded $D,s$ and fixed $m>0$, $u^2\sim m/a$ and $g\sim2am$. The gain grows from $O(a^2)$ to $O(a)$ as the factors learn. A weak residual shared by all replicas can therefore become much more influential than its initial response suggests.

\subsection{A phase determined by raw initialization}
In slow time $\tau=at_{\rm GF}$, the normalized limiting orbit solves
\begin{equation}
 M'=2M[c-H(M)],\qquad
 \lim_{t\to-\infty}e^{-2ct}M(t)=1.
 \label{eq:mainorbit}
\end{equation}
It increases from zero to $m_*=[\operatorname{logit}(q_+)-\operatorname{logit}(q_-)]/2$, with $s_*=[\operatorname{logit}(q_+)+\operatorname{logit}(q_-)]/2$. Define the phase from raw weights $\theta_0=(u_0,v_{10},v_{20})$. Set $D_0=u_0^2-v_{10}^2-v_{20}^2$, $w_0=\operatorname{sign}(u_0+v_{10})v_{20}$, and let $\Gamma_D(w)>0$ solve $\Gamma_D(w)^2-s_0^2/\Gamma_D(w)^2=D+w^2$. Then
\begin{equation}
 \begin{aligned}
 \beta(\theta_0)&=\lim_{R\to\infty}R^2
   \exp\!\left[-2\int_{w_0}^R\frac{\dd w}{\Gamma_{D_0}(w)}\right],\\
 \chi(\theta_0)&=\frac{\log\beta(\theta_0)}{2c}.
 \end{aligned}\label{eq:mainphase}
\end{equation}
The sign fixes a function-preserving gauge. This phase is computed from the raw weights before training.

\paragraph{How initialization survives strong alignment.}
The fast fit constrains the product $uv_1$ while retaining information in the factor balance and the weak weight. At $a=0$, $v_2$ is fixed, and the sum and difference of the strong factors satisfy $\dot{(u+v_1)}=A(u+v_1)$ and $\dot{(u-v_1)}=-A(u-v_1)$. Outside the Gaussian-null line $u_0+v_{10}=0$, the product approaches $s_0$, and the two factors acquire the common sign fixed by the raw initialization. For small positive $a$, an initial interval of order $\log(1/a)$ brings the strong tracking error to $O(a)$ while changing the weak weight by $o(1)$. The replicas can therefore reach nearly the same strong predictions with different weak starting states.

In the positive-$u$ gauge, balance and strong tracking give $u=\Gamma_{D_0}(w)+o(1)$ on bounded weak-weight intervals. For fixed large $R>w_0$, the slow time to reach $R$ is consequently governed by
\begin{equation}
 \tau_R=\frac1c\int_{w_0}^R\frac{\dd w}{\Gamma_{D_0}(w)}+o(1).
 \label{eq:maininnertime}
\end{equation}
Both the lower endpoint and the gain in this integral depend on initialization. Since $\Gamma_D(w)\sim w$, the remaining growth becomes exponential and preserves that dependence as the amplitude $\beta$. Matching $m\sim a\beta e^{2c\tau}$ to the outer orbit converts the amplitude into the time shift $\chi$. This identifies where the replicas retain their differences during early functional alignment and how those differences reappear during weak learning. The limiting orbit is shared, while each replica reaches its transition at its own initialization-dependent phase.

\begin{theorem}[Initialization-dependent phases]\label{thm:twophase}
Condition on the fixed dataset above and let $\tau_a=\log(1/a)/(2c)$. For gradient flow, almost surely over independent Gaussian initializations, uniformly on compact intervals of shifted time as $a\to0$,
\begin{equation}
 \begin{aligned}
 m_{j,a}\big((\tau_a+t)/a\big)&\longrightarrow M(t+\chi_j),\\
 s_{j,a}\big((\tau_a+t)/a\big)&\longrightarrow s^*(M(t+\chi_j)),\\
 \chi_j&=\chi(\theta_{j0}).
 \end{aligned}
 \label{eq:twophase}
\end{equation}
For sufficiently small fixed $a$, the effective logits also converge to $(s_*,m_*)$ as training time tends to infinity.
\end{theorem}

The theorem formalizes this matching argument. Full statements and proofs are in \cref{sec:twolayerdetails,sec:twoproofs}. Stable full-batch GD has the same phase limit on compact initialization events away from the Gaussian-null unstable boundary when $0<\eta\le\nu a$, for an event-dependent $\nu>0$, evaluated at $k=\lfloor(\tau_a+t)/(\eta a)\rfloor$.

On the conditional strong curve, symmetric KL simplifies to
\begin{equation}
 \mathcal S(m_0,m_1)=\tfrac12(m_0-m_1)[H(m_0)-H(m_1)].
 \label{eq:mainKL}
\end{equation}
Independent Gaussian initializations give unequal phases almost surely. They agree near the beginning of the orbit, separate when traversing it at different times, and agree again at its common endpoint. The following consequence makes the three-stage comparison precise.

\begin{corollary}[Three-stage replica discrepancy]\label{cor:twothree}
For every $\delta\in(0,1)$ and $\epsilon>0$, there are deterministic $t_1<0=t_2<t_3$, a gap $g_0>0$, and $a_0>0$ such that for $a<a_0$, with probability at least $1-\delta$ over the two initializations,
\begin{equation}
 \begin{gathered}
 S_a(0)>S_a^{[1]},\qquad S_a^{[2]}-S_a^{[1]}\ge g_0,\\
 S_a^{[3]}<\min(\epsilon,g_0),
 \end{gathered}
 \label{eq:twothree}
\end{equation}
where $S_a^{[\ell]}$ is evaluated at $(\tau_a+t_\ell)/a$. For each sufficiently small fixed $a$, $S_a(t)\to0$ as $t\to\infty$. The stable-GD version uses the discrete times above.
\end{corollary}

\paragraph{Why the phase limit gives a resurgence.}
On the conditional curve, group probabilities have the same mean $\bar q$ in both replicas. In the Bernoulli KL identity, the strong-logit difference therefore cancels, leaving \cref{eq:mainKL}. Strict monotonicity of $H$ makes this expression positive whenever the two phases give different $m$. At either end of the orbit, both coefficients approach the same endpoint and the expression vanishes. Quantiles of the random phases provide deterministic comparison times and a positive middle gap with the stated high probability.

\begin{figure*}[t]\centering
\includegraphics[width=\linewidth]{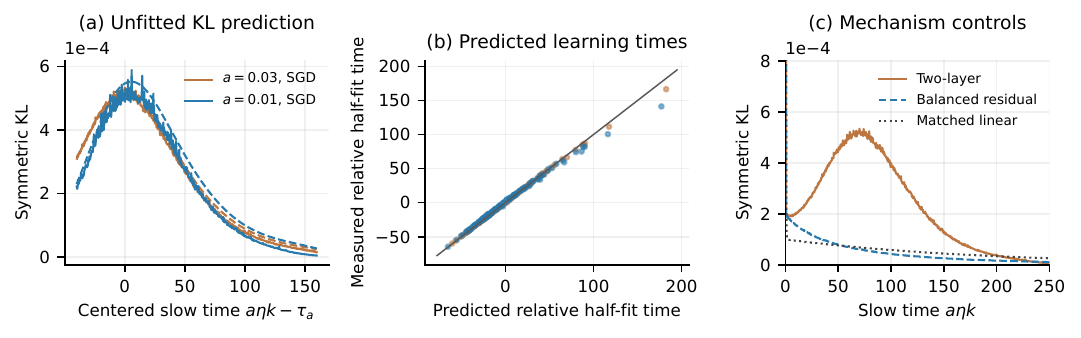}
\caption{\textbf{Tests of the two-layer mechanism.} (a) Measured SGD KL (solid) and finite-$a$ reduction (dashed), computed directly from data and raw initialization. (b) Predicted versus measured within-stream relative half-fit times. (c) Balancing the weak residual and replacing the network by a single-layer predictor with the same initial function remove the comparable resurgence. The vertical range in (c) resolves the post-alignment phase. Full trajectories are in the supplement.}\label{fig:maintheory}
\end{figure*}

\subsection{Discrepancy and population risk}
We next examine what the phase mechanism implies for prediction quality. We compare tasks in which the learned weak direction carries spurious correlations or genuine predictive information and derive the corresponding population-risk trends.

When the true label is $Y=b$, the weak dependence in \cref{eq:mainloss} is spurious. On the conditional curve its clean population CE is
\begin{equation}
 \begin{gathered}
 \mathcal R(m)=-\tfrac12\log[\bar q^2-H(m)^2],\\
 \mathcal R'(m)>0\quad(m>0).
 \end{gathered}
 \label{eq:mainrisk}
\end{equation}
Thus replica discrepancy ultimately vanishes while clean CE increases by $-\tfrac12\log(1-c^2/\bar q^2)>0$. Both replicas acquire the same harmful dependence as their predictions converge.

When $q_->1/2$, both group probabilities $\bar q\pm H(m)$ exceed $1/2$ throughout the limiting orbit, since $0\le H(m)\le c$. Clean classification error then stays zero during weak learning, while population CE increases and replica KL resurges. The phase mechanism therefore separates the evolution of prediction differences from that of classification error even within one task.

The same phase mechanism also accommodates improving risk. Keep the same trainable network and inputs, let $\PP(t=+1)=\pi>1/2$, and set observed and true labels deterministically to $Y=bt$. The strong coordinate is now a shortcut of accuracy $\pi$, while the weak coordinate is perfectly predictive. The loss is
\begin{equation}
 \mathcal L_{\rm cl}=\pi\log(1+e^{-s-m})+(1-\pi)\log(1+e^{s-m}).
\end{equation}
On its conditional strong curve, let $J(m)=\pi[1-\sigma(s^*_{\rm cl}(m)+m)]=(1-\pi)\sigma(s^*_{\rm cl}(m)-m)$. The same GF matching argument gives $M'_{\rm cl}=4M_{\rm cl}J(M_{\rm cl})$ and raw-initialization phases. Now $\mathcal R'_{\rm cl}(m)=-2J(m)<0$, while $\mathcal S_{\rm cl}=-(m_0-m_1)[J(m_0)-J(m_1)]$ again vanishes at both ends of the limiting phase window. This extension concerns GF on finite shifted-time windows (\cref{sec:cleanproof}). Across the two instantiations, phase differences govern discrepancy. Task information governs risk.

\subsection{Quantitative tests and mechanism controls}

For a threshold $m_{\rm cut}\in(0,m_*)$, the theorem predicts the crossing time $T_{j,a}$ through
\begin{equation}
 aT_{j,a}-\tau_a\longrightarrow M^{-1}(m_{\rm cut})-\chi_j.
\end{equation}
For numerical prediction at finite $a$, we retain the exact balance relation and conditional strong tracking throughout training. All inputs come from raw weights and the dataset (\cref{sec:twoprediction}).

In particular, differentiating the conditional relation $s=s^*(m)$ and eliminating $A$ from \cref{eq:mainkernel} gives the reduced equation
\begin{equation}
 \frac{\dd m}{\dd\tau}
 =\frac{K_{ss}K_{mm}-K_{sm}^2}
 {a[K_{ss}-s^{*\prime}(m)K_{sm}]}[c-H(m)].
 \label{eq:mainfinitereduction}
\end{equation}
The initial state is determined by the fast strong fit. Put $C_0=u_0^2-v_{10}^2$. At leading order, this fit sets $uv_1=s_0$ while leaving $v_2=v_{20}$. In this fast limit, conservation of $u^2-v_1^2=C_0$ gives
\begin{equation}
 \begin{aligned}
 U_{\rm init}&=\tfrac12[C_0+\sqrt{C_0^2+4s_0^2}],\\
 m_{\rm init}&=a\operatorname{sign}(u_0+v_{10})\sqrt{U_{\rm init}}\,v_{20}.
 \end{aligned}\label{eq:mainprojectedinit}
\end{equation}
Here $U_{\rm init}$ is the squared output weight after strong fitting. We initialize \cref{eq:mainfinitereduction} with $m_{\rm init}$ and evaluate $K$ using the raw balance $D_0$ and \cref{eq:mainbalance}. These initialization-dependent quantities determine the predicted learning times (\cref{sec:twoprediction,sec:twonumerics}).

Taking the difference of two predicted crossing times eliminates the threshold-dependent origin.
\begin{equation}
 a(T_{0,a}-T_{1,a})\longrightarrow\chi_1-\chi_0.
 \label{eq:mainrelativetime}
\end{equation}
Thus raw initialization predicts the ordering and relative delay of threshold crossings. The common transition occurs near physical time $\log(1/a)/(2ac)$, with phase offsets visible on the $1/a$ time scale.

We test the predicted trajectory and learning times in the two-layer linear network on the fixed binary task with 10\% label flips, using full-batch GD and shared-minibatch SGD. In \cref{fig:maintheory}(a), the finite-$a$ reduction follows the SGD resurgence and subsequent decline after initial alignment. Panel (b) shows that predicted relative half-fit times lie close to the measured values across initializations. Thus the raw weights predict both the ordering and separation of later learning times. The full trajectories and evolving weak gain in \cref{fig:twomechanism} connect this timing variation to the amplification of the weak response. Training configurations and prediction errors appear in \cref{sec:twonumerics}.

Balancing group targets sets $c=0$ and removes the weak instability. Indeed, GF then satisfies $\dd(v_2^2)/\dd t=-2m h(s,m)\le0$, where $h(s,m)=[\sigma(s+m)-\sigma(s-m)]/2$. This sign holds before imposing strong tracking. The balanced control therefore removes the residual driving weak amplification.

A separate single-layer predictor starts at coefficients $(u_0v_{10},u_0v_{20})$, exactly matching each replica's initial function. Its weak gain is the constant $a^2$, whereas the two-layer gain evolves according to \cref{eq:maingain}. In \cref{fig:maintheory}(c), the two-layer replicas separate while both controls remain approximately aligned over the tested post-alignment interval. Together, these interventions connect the resurgence to a shared weak residual acting through an amplifying learned response.

The input scale controls how clearly the stages separate. At $a=0.1$, strong and weak learning overlap and no distinct rise appears (\cref{fig:twomechanism}). In the predictive task, \cref{fig:cleanphasevalidation} shows an intermediate KL resurgence while clean population CE decreases, as the replicas learn the predictive weak direction. Raw phase timing is accurate at $a=0.003$ and substantially less accurate at $0.01$ and $0.03$. The supplement reports timing errors across these scales.

\subsection{Stochastic corrections}
Finite-step SGD introduces a second-order correction to the balance used in the phase proof. For the empirical task, define residuals $r_i=T_i-\sigma(s+t_im)$, and batch averages $A_{\mathcal B}=\operatorname{mean}_{i\in\mathcal B}r_i$, $B_{\mathcal B}=\operatorname{mean}_{i\in\mathcal B}t_ir_i$. A direct expansion of the three factor updates gives
\begin{equation}
 \begin{aligned}
 D^+-D=\eta^2\big[&(v_1A_{\mathcal B}+av_2B_{\mathcal B})^2\\
 &-u^2(A_{\mathcal B}^2+a^2B_{\mathcal B}^2)\big].
 \end{aligned}\label{eq:mainSGDbalance}
\end{equation}
The terms linear in $\eta$ cancel, leaving the displayed quadratic correction. Its conditional expectation combines mean-gradient and gradient-variance contributions in a signed difference. Both contributions depend on the current factors, so replicas can accumulate different balance corrections under the same sampled examples.

To examine the effect of these corrections, we vary minibatch size in the two-layer linear-network task with 10\% label flips. At $a=0.03$, reducing the batch size from 32 to 1 produces a smaller discrepancy resurgence (\cref{fig:twocontrols}). The accompanying numerical audit shows greater balance drift before half-fitting and less accurate relative-time predictions (\cref{sec:twonumerics}). Over the long weak-learning interval, the per-step corrections accumulate and change the balance that enters the timing prediction. Tracking this drift therefore helps diagnose departures from the GF-derived learning times under SGD.

The model links strong alignment, weak-response amplification, and initialization-dependent timing. Together with the CNN interventions and sample histories, it motivates tracking both current agreement and future update responses.

\section{Discussion and Conclusion}\label{sec:applications}
Networks trained on identical data can become more different after an initial period of alignment. Controlled vision experiments document this temporal resurgence when initialization is the only difference between runs. Sample-level analyses connect the aggregate trajectory to changing prediction contributions and asymmetric forgetting. A fully trained two-layer model derives the resurgence from initialization-dependent phases of an amplified weak response. The same phase mechanism permits improving or worsening population risk, depending on what the replicas learn in common.

These findings suggest evaluating predictive reproducibility throughout training. A monitor could combine held-out discrepancy, risk, and sample contributions to track prediction quality and the examples driving replica differences. Ensemble diversity supports predictive uncertainty estimation \cite{lakshminarayanan2017,fort2019}. Evaluating ensembles across learning phases could track changes in this diversity. Tracking current agreement and shared-update responses would help identify when training suppresses or amplifies initialization-dependent variation.

\clearpage
\section*{LLM Usage Statement}
Codex was used for literature search, manuscript editing, development and checking of theoretical derivations, and implementation of experiments. The reported claims, proofs, references, and results remain the responsibility of the authors.

\begingroup
\small
\bibliographystyle{ieeenat_fullname_unsrt}
\bibliography{references}

@inproceedings{jiang2022,
 title={Assessing Generalization of {SGD} via Disagreement}, author={Jiang, Yiding and Nagarajan, Vaishnavh and Baek, Christina and Kolter, J. Zico},
 booktitle={International Conference on Learning Representations}, year={2022}, url={https://arxiv.org/abs/2106.13799}
}

@article{kirsch2022,
 title={A Note on ``Assessing Generalization of {SGD} via Disagreement''}, author={Kirsch, Andreas and Gal, Yarin},
 journal={Transactions on Machine Learning Research}, year={2022}, url={https://arxiv.org/abs/2202.01851}
}

@inproceedings{somepalli2022,
 title={Can Neural Nets Learn the Same Model Twice? Investigating Reproducibility and Double Descent From the Decision Boundary Perspective},
 author={Somepalli, Gowthami and Fowl, Liam and Bansal, Arpit and Yeh-Chiang, Ping and Dar, Yehuda and Baraniuk, Richard and Goldblum, Micah and Goldstein, Tom},
 booktitle={Proceedings of the IEEE/CVF Conference on Computer Vision and Pattern Recognition}, pages={13699--13708}, year={2022}, url={https://arxiv.org/abs/2203.08124}
}

@inproceedings{adlam2020,
 title={Understanding Double Descent Requires a Fine-Grained Bias-Variance Decomposition}, author={Adlam, Ben and Pennington, Jeffrey},
 booktitle={Advances in Neural Information Processing Systems}, volume={33}, year={2020}, url={https://arxiv.org/abs/2011.03321}
}

@inproceedings{dascoli2020,
 title={Double Trouble in Double Descent: Bias and Variance(s) in the Lazy Regime},
 author={d'Ascoli, St{\'e}phane and Refinetti, Maria and Biroli, Giulio and Krzakala, Florent},
 booktitle={Proceedings of the 37th International Conference on Machine Learning},
 volume={119}, pages={2280--2290}, year={2020}, url={https://proceedings.mlr.press/v119/d-ascoli20a.html}
}

@inproceedings{liu2022,
 title={On the Double Descent of Random Features Models Trained with {SGD}}, author={Liu, Fanghui and Suykens, Johan A. K. and Cevher, Volkan},
 booktitle={Advances in Neural Information Processing Systems}, volume={35}, year={2022}, url={https://arxiv.org/abs/2110.06910}
}

@article{atanasov2025,
 title={Two-point deterministic equivalence for stochastic gradient dynamics in linear models},
 author={Atanasov, Alexander and Bordelon, Blake and Zavatone-Veth, Jacob A. and Paquette, Courtney and Pehlevan, Cengiz},
 journal={Advances in Theoretical and Mathematical Physics}, volume={30}, number={1}, year={2026}, doi={10.4310/ATMP.260412235902}, url={https://arxiv.org/abs/2502.05074}
}

@article{bordelon2026,
 title={Disordered Dynamics in High Dimensions: Connections to Random Matrices and Machine Learning}, author={Bordelon, Blake and Pehlevan, Cengiz},
 journal={arXiv preprint arXiv:2601.01010}, year={2026}, url={https://arxiv.org/abs/2601.01010}
}

@inproceedings{han2025,
 title={On the Role of Label Noise in the Feature Learning Process},
 author={Han, Andi and Huang, Wei and Zhou, Zhanpeng and Niu, Gang and Chen, Wuyang and Yan, Junchi and Takeda, Akiko and Suzuki, Taiji},
 booktitle={Proceedings of the 42nd International Conference on Machine Learning}, volume={267}, pages={21733--21773}, year={2025}, url={https://proceedings.mlr.press/v267/han25c.html}
}

@inproceedings{wei2022,
 title={Learning with Noisy Labels Revisited: A Study Using Real-World Human Annotations},
 author={Wei, Jiaheng and Zhu, Zhaowei and Cheng, Hao and Liu, Tongliang and Niu, Gang and Liu, Yang},
 booktitle={International Conference on Learning Representations}, year={2022}, url={https://arxiv.org/abs/2110.12088}
}

@inproceedings{he2016,
 title={Deep Residual Learning for Image Recognition}, author={He, Kaiming and Zhang, Xiangyu and Ren, Shaoqing and Sun, Jian},
 booktitle={Proceedings of the IEEE Conference on Computer Vision and Pattern Recognition}, pages={770--778}, year={2016}, url={https://openaccess.thecvf.com/content_cvpr_2016/html/He_Deep_Residual_Learning_CVPR_2016_paper.html}
}

@inproceedings{loshchilov2019,
 title={Decoupled Weight Decay Regularization}, author={Loshchilov, Ilya and Hutter, Frank},
 booktitle={International Conference on Learning Representations}, year={2019}, url={https://arxiv.org/abs/1711.05101}
}

@inproceedings{touvron2021,
 title={Training data-efficient image transformers \& distillation through attention},
 author={Touvron, Hugo and Cord, Matthieu and Douze, Matthijs and Massa, Francisco and Sablayrolles, Alexandre and J{\'e}gou, Herv{\'e}},
 booktitle={Proceedings of the 38th International Conference on Machine Learning}, volume={139}, pages={10347--10357}, year={2021}, url={https://proceedings.mlr.press/v139/touvron21a.html}
}

@inproceedings{even2023,
 title={(S)GD over Diagonal Linear Networks: Implicit Regularisation, Large Stepsizes and Edge of Stability},
 author={Even, Mathieu and Pesme, Scott and Gunasekar, Suriya and Flammarion, Nicolas},
 booktitle={Advances in Neural Information Processing Systems}, volume={36}, year={2023}, url={https://arxiv.org/abs/2302.08982}
}

@inproceedings{saxe2014,
 title={Exact solutions to the nonlinear dynamics of learning in deep linear neural networks},
 author={Saxe, Andrew M. and McClelland, James L. and Ganguli, Surya},
 booktitle={International Conference on Learning Representations}, year={2014}, url={https://arxiv.org/abs/1312.6120}
}

@inproceedings{pesme2023,
 title={Saddle-to-Saddle Dynamics in Diagonal Linear Networks}, author={Pesme, Scott and Flammarion, Nicolas},
 booktitle={Advances in Neural Information Processing Systems}, volume={36}, year={2023},
 url={https://papers.neurips.cc/paper_files/paper/2023/hash/17a9ab4190289f0e1504bbb98d1d111a-Abstract-Conference.html}
}

@inproceedings{nakkiran2020,
 title={Deep Double Descent: Where Bigger Models and More Data Hurt},
 author={Nakkiran, Preetum and Kaplun, Gal and Bansal, Yamini and Yang, Tristan and Barak, Boaz and Sutskever, Ilya},
 booktitle={International Conference on Learning Representations}, year={2020},
 url={https://arxiv.org/abs/1912.02292}
}

@inproceedings{wu2018,
 title={Group Normalization}, author={Wu, Yuxin and He, Kaiming},
 booktitle={Proceedings of the European Conference on Computer Vision},
 pages={3--19}, year={2018},
 url={https://openaccess.thecvf.com/content_ECCV_2018/html/Yuxin_Wu_Group_Normalization_ECCV_2018_paper.html}
}

@inproceedings{atanasov2022,
 title={Neural Networks as Kernel Learners: The Silent Alignment Effect},
 author={Atanasov, Alexander and Bordelon, Blake and Pehlevan, Cengiz},
 year={2022},
 booktitle={International Conference on Learning Representations},
 url={https://openreview.net/forum?id=1NvflqAdoom}
}

@article{martin2026,
 title={Silent Alignment in Two-Layer Linear Networks: A High-Dimensional Perspective},
 author={Martin, Simon and Bach, Francis and Biroli, Giulio},
 year={2026},
 journal={arXiv preprint arXiv:2610.29864},
 url={https://arxiv.org/abs/2610.29864}
}

@inproceedings{zhang2026,
 title={Saddle-to-Saddle Dynamics Explains a Simplicity Bias Across Neural Network Architectures},
 author={Zhang, Yedi and Saxe, Andrew and Latham, Peter E.},
 year={2026},
 booktitle={International Conference on Learning Representations},
 url={https://arxiv.org/abs/2512.20607}
}

@inproceedings{kunin2024,
 title={Get rich quick: exact solutions reveal how unbalanced initializations promote rapid feature learning},
 author={Kunin, Daniel and Ravent{\'o}s, Allan and Domin{\'e}, Cl{\'e}mentine and Chen, Feng and Klindt, David and Saxe, Andrew and Ganguli, Surya},
 year={2024},
 booktitle={Advances in Neural Information Processing Systems},
 volume={37},
 url={https://proceedings.neurips.cc/paper_files/paper/2024/hash/94074dd5a072d28ff75a76dabed43767-Abstract-Conference.html}
}

@inproceedings{xu2025,
 title={Three Mechanisms of Feature Learning in a Linear Network},
 author={Xu, Yizhou and {Liu Ziyin}},
 year={2025},
 booktitle={International Conference on Learning Representations},
 url={https://proceedings.iclr.cc/paper_files/paper/2025/hash/9941833e8327910ef25daeb9005e4748-Abstract-Conference.html}
}

@inproceedings{du2018,
 title={Algorithmic Regularization in Learning Deep Homogeneous Models: Layers are Automatically Balanced},
 author={Du, Simon S. and Hu, Wei and Lee, Jason D.},
 year={2018},
 booktitle={Advances in Neural Information Processing Systems},
 volume={31},
 url={https://proceedings.neurips.cc/paper_files/paper/2018/hash/fe131d7f5a6b38b23cc967316c13dae2-Abstract.html}
}

@inproceedings{chizat2019,
 title={On Lazy Training in Differentiable Programming},
 author={Chizat, L{\'e}na{\"i}c and Oyallon, Edouard and Bach, Francis},
 year={2019},
 booktitle={Advances in Neural Information Processing Systems},
 volume={32},
 url={https://proceedings.neurips.cc/paper/2019/hash/ae614c557843b1df326cb29c57225459-Abstract.html}
}

@inproceedings{arpit2017,
 title={A Closer Look at Memorization in Deep Networks},
 author={Arpit, Devansh and Jastrz{\k{e}}bski, Stanis{\l}aw and Ballas, Nicolas and Krueger, David and Bengio, Emmanuel and Kanwal, Maxinder S. and Maharaj, Tegan and Fischer, Asja and Courville, Aaron and Bengio, Yoshua and Lacoste-Julien, Simon},
 year={2017},
 booktitle={Proceedings of the 34th International Conference on Machine Learning},
 volume={70},
 pages={233--242},
 url={https://proceedings.mlr.press/v70/arpit17a.html}
}

@article{power2022,
 title={Grokking: Generalization Beyond Overfitting on Small Algorithmic Datasets},
 author={Power, Alethea and Burda, Yuri and Edwards, Harri and Babuschkin, Igor and Misra, Vedant},
 year={2022},
 journal={arXiv preprint arXiv:2201.02177},
 url={https://arxiv.org/abs/2201.02177}
}

@inproceedings{kumar2024,
 title={Grokking as the Transition from Lazy to Rich Training Dynamics},
 author={Kumar, Tanishq and Bordelon, Blake and Gershman, Samuel J. and Pehlevan, Cengiz},
 year={2024},
 booktitle={International Conference on Learning Representations},
 url={https://openreview.net/forum?id=vt5mnLVIVo}
}

@article{belkin2019,
 title={Reconciling modern machine-learning practice and the classical bias--variance trade-off},
 author={Belkin, Mikhail and Hsu, Daniel and Ma, Siyuan and Mandal, Soumik},
 year={2019},
 journal={Proceedings of the National Academy of Sciences},
 volume={116},
 number={32},
 pages={15849--15854},
 url={https://doi.org/10.1073/pnas.1903070116}
}

@inproceedings{heckel2021,
 title={Early Stopping in Deep Networks: Double Descent and How to Eliminate it},
 author={Heckel, Reinhard and Yilmaz, Fatih Furkan},
 year={2021},
 booktitle={International Conference on Learning Representations},
 url={https://openreview.net/forum?id=tlV90jvZbw}
}

@inproceedings{pezeshki2022,
 title={Multi-scale Feature Learning Dynamics: Insights for Double Descent},
 author={Pezeshki, Mohammad and Mitra, Amartya and Bengio, Yoshua and Lajoie, Guillaume},
 year={2022},
 booktitle={Proceedings of the 39th International Conference on Machine Learning},
 volume={162},
 pages={17669--17690},
 url={https://proceedings.mlr.press/v162/pezeshki22a.html}
}

@article{stephenson2021,
 title={When and how epochwise double descent happens},
 author={Stephenson, Cory and Lee, Tyler},
 year={2021},
 journal={arXiv preprint arXiv:2108.12006},
 url={https://arxiv.org/abs/2108.12006}
}

@inproceedings{summers2021,
 title={Nondeterminism and Instability in Neural Network Optimization},
 author={Summers, Cecilia and Dinneen, Michael J.},
 year={2021},
 booktitle={Proceedings of the 38th International Conference on Machine Learning},
 volume={139},
 pages={9913--9922},
 url={https://proceedings.mlr.press/v139/summers21a.html}
}

@article{bhojanapalli2021,
 title={On the Reproducibility of Neural Network Predictions},
 author={Bhojanapalli, Srinadh and Wilber, Kimberly and Veit, Andreas and Rawat, Ankit Singh and Kim, Seungyeon and Menon, Aditya and Kumar, Sanjiv},
 year={2021},
 journal={arXiv preprint arXiv:2102.03349},
 url={https://arxiv.org/abs/2102.03349}
}

@inproceedings{bahri2021,
 title={Locally Adaptive Label Smoothing Improves Predictive Churn},
 author={Bahri, Dara and Jiang, Heinrich},
 year={2021},
 booktitle={Proceedings of the 38th International Conference on Machine Learning},
 volume={139},
 pages={532--542},
 url={https://proceedings.mlr.press/v139/bahri21a.html}
}

@article{adam2023,
 title={Maintaining Stability and Plasticity for Predictive Churn Reduction},
 author={Adam, George and Haibe-Kains, Benjamin and Goldenberg, Anna},
 year={2023},
 journal={arXiv preprint arXiv:2305.04135},
 url={https://arxiv.org/abs/2305.04135}
}

@article{damour2022,
 title={Underspecification Presents Challenges for Credibility in Modern Machine Learning},
 author={D'Amour, Alexander and Heller, Katherine and Moldovan, Dan and Adlam, Ben and Alipanahi, Babak and Beutel, Alex and Chen, Christina and Deaton, Jonathan and Eisenstein, Jacob and Hoffman, Matthew D. and Hormozdiari, Farhad and Houlsby, Neil and Hou, Shaobo and Jerfel, Ghassen and Karthikesalingam, Alan and Lucic, Mario and Ma, Yian and McLean, Cory and Mincu, Diana and Mitani, Akinori and Montanari, Andrea and Nado, Zachary and Natarajan, Vivek and Nielson, Christopher and Osborne, Thomas F. and Raman, Rajiv and Ramasamy, Kim and Sayres, Rory and Schrouff, Jessica and Seneviratne, Martin and Sequeira, Shannon and Suresh, Harini and Veitch, Victor and Vladymyrov, Max and Wang, Xuezhi and Webster, Kellie and Yadlowsky, Steve and Yun, Taedong and Zhai, Xiaohua and Sculley, D.},
 year={2022},
 journal={Journal of Machine Learning Research},
 volume={23},
 number={226},
 pages={1--61},
 url={https://www.jmlr.org/papers/v23/20-1335.html}
}

@inproceedings{frankle2020,
 title={Linear Mode Connectivity and the Lottery Ticket Hypothesis},
 author={Frankle, Jonathan and Dziugaite, Gintare Karolina and Roy, Daniel and Carbin, Michael},
 year={2020},
 booktitle={Proceedings of the 37th International Conference on Machine Learning},
 volume={119},
 pages={3259--3269},
 url={https://proceedings.mlr.press/v119/frankle20a.html}
}

@inproceedings{fort2020,
 title={Deep learning versus kernel learning: an empirical study of loss landscape geometry and the time evolution of the Neural Tangent Kernel},
 author={Fort, Stanislav and Dziugaite, Gintare Karolina and Paul, Mansheej and Kharaghani, Sepideh and Roy, Daniel M. and Ganguli, Surya},
 year={2020},
 booktitle={Advances in Neural Information Processing Systems},
 volume={33},
 url={https://proceedings.neurips.cc/paper/2020/hash/405075699f065e43581f27d67bb68478-Abstract.html}
}

@inproceedings{butterfly2025,
 title={The Butterfly Effect: Neural Network Training Trajectories Are Highly Sensitive to Initial Conditions},
 author={Alt{\i}nta\c{s}, G{\"u}l Sena and Kwok, Devin and Raffel, Colin and Rolnick, David},
 year={2025},
 booktitle={Proceedings of the 42nd International Conference on Machine Learning},
 volume={267},
 pages={1314--1342},
 url={https://proceedings.mlr.press/v267/altintas25a.html}
}

@article{ly2026,
 title={Fractal basin geometry and the limits to predictability and reproducibility in deep learning},
 author={Ly, Andrew and Gong, Pulin},
 year={2026},
 journal={Nature Machine Intelligence},
 doi={10.1038/s42256-026-01336-3},
 url={https://www.nature.com/articles/s42256-026-01336-3}
}

@inproceedings{baek2022,
 title={Agreement-on-the-line: Predicting the Performance of Neural Networks under Distribution Shift},
 author={Baek, Christina and Jiang, Yiding and Raghunathan, Aditi and Kolter, J. Zico},
 year={2022},
 booktitle={Advances in Neural Information Processing Systems},
 volume={35},
 url={https://proceedings.neurips.cc/paper_files/paper/2022/hash/7a8d388b7a17df480856dff1cc079b08-Abstract-Conference.html}
}

@inproceedings{lee2023,
 title={Demystifying Disagreement-on-the-Line in High Dimensions},
 author={Lee, Donghwan and Moniri, Behrad and Huang, Xinmeng and Dobriban, Edgar and Hassani, Hamed},
 year={2023},
 booktitle={Proceedings of the 40th International Conference on Machine Learning},
 volume={202},
 pages={19053--19093},
 url={https://proceedings.mlr.press/v202/lee23o.html}
}

@inproceedings{saxena2024,
 title={Predicting the Performance of Foundation Models via Agreement-on-the-Line},
 author={Saxena, Rahul and Kim, Taeyoun and Mehra, Aman and Baek, Christina and Kolter, Zico and Raghunathan, Aditi},
 year={2024},
 booktitle={Advances in Neural Information Processing Systems},
 volume={37},
 url={https://proceedings.neurips.cc/paper_files/paper/2024/hash/3853f7b92d82318f03567a4f7f128227-Abstract-Conference.html}
}

@inproceedings{eaton2026,
 title={Model Agreement via Anchoring},
 author={Eaton, Eric and Goel, Surbhi and Hussing, Marcel and Kearns, Michael and Roth, Aaron and Sengupta, Sikata Bela and Sorrell, Jessica},
 year={2026},
 booktitle={Proceedings of Thirty Ninth Conference on Learning Theory},
 volume={336},
 pages={5625--5661},
 url={https://proceedings.mlr.press/v336/eaton26a.html}
}

@inproceedings{toneva2019,
 title={An Empirical Study of Example Forgetting during Deep Neural Network Learning},
 author={Toneva, Mariya and Sordoni, Alessandro and {Tachet des Combes}, Remi and Trischler, Adam and Bengio, Yoshua and Gordon, Geoffrey J.},
 year={2019},
 booktitle={International Conference on Learning Representations},
 url={https://openreview.net/forum?id=BJlxm30cKm}
}

@inproceedings{jacot2018,
 title={Neural Tangent Kernel: Convergence and Generalization in Neural Networks},
 author={Jacot, Arthur and Gabriel, Franck and Hongler, Cl{\'e}ment},
 year={2018},
 booktitle={Advances in Neural Information Processing Systems},
 volume={31},
 url={https://proceedings.neurips.cc/paper/2018/hash/5a4be1fa34e62bb8a6ec6b91d2462f5a-Abstract.html}
}

@article{martens2020,
 title={New Insights and Perspectives on the Natural Gradient Method},
 author={Martens, James},
 year={2020},
 journal={Journal of Machine Learning Research},
 volume={21},
 number={146},
 pages={1--76},
 url={https://jmlr.org/papers/v21/17-678.html}
}

@inproceedings{dinh2017,
 title={Sharp Minima Can Generalize For Deep Nets},
 author={Dinh, Laurent and Pascanu, Razvan and Bengio, Samy and Bengio, Yoshua},
 year={2017},
 booktitle={Proceedings of the 34th International Conference on Machine Learning},
 volume={70},
 pages={1019--1028},
 url={https://proceedings.mlr.press/v70/dinh17b.html}
}

@techreport{krizhevsky2009,
 title={Learning Multiple Layers of Features from Tiny Images},
 author={Krizhevsky, Alex},
 year={2009},
 institution={University of Toronto},
 url={https://www.cs.toronto.edu/~kriz/learning-features-2009-TR.pdf}
}

@inproceedings{deng2009,
 title={{ImageNet}: A Large-Scale Hierarchical Image Database},
 author={Deng, Jia and Dong, Wei and Socher, Richard and Li, Li-Jia and Li, Kai and Fei-Fei, Li},
 year={2009},
 booktitle={Proceedings of the IEEE Conference on Computer Vision and Pattern Recognition},
 pages={248--255},
 url={https://www.image-net.org/static_files/papers/imagenet_cvpr09.pdf}
}

@inproceedings{tian2020,
 title={Contrastive Multiview Coding},
 author={Tian, Yonglong and Krishnan, Dilip and Isola, Phillip},
 year={2020},
 booktitle={Proceedings of the European Conference on Computer Vision},
 url={https://www.ecva.net/papers/eccv_2020/papers_ECCV/html/1453_ECCV_2020_paper.php}
}

@inproceedings{dosovitskiy2021,
 title={An Image is Worth 16x16 Words: Transformers for Image Recognition at Scale},
 author={Dosovitskiy, Alexey and Beyer, Lucas and Kolesnikov, Alexander and Weissenborn, Dirk and Zhai, Xiaohua and Unterthiner, Thomas and Dehghani, Mostafa and Minderer, Matthias and Heigold, Georg and Gelly, Sylvain and Uszkoreit, Jakob and Houlsby, Neil},
 year={2021},
 booktitle={International Conference on Learning Representations},
 url={https://arxiv.org/abs/2010.11929}
}

@inproceedings{glorot2010,
 title={Understanding the difficulty of training deep feedforward neural networks},
 author={Glorot, Xavier and Bengio, Yoshua},
 year={2010},
 booktitle={Proceedings of the Thirteenth International Conference on Artificial Intelligence and Statistics},
 volume={9},
 pages={249--256},
 url={https://proceedings.mlr.press/v9/glorot10a.html}
}

@inproceedings{he2015,
 title={Delving Deep into Rectifiers: Surpassing Human-Level Performance on {ImageNet} Classification},
 author={He, Kaiming and Zhang, Xiangyu and Ren, Shaoqing and Sun, Jian},
 year={2015},
 booktitle={Proceedings of the IEEE International Conference on Computer Vision},
 pages={1026--1034},
 url={https://openaccess.thecvf.com/content_iccv_2015/html/He_Delving_Deep_into_ICCV_2015_paper.html}
}

@inproceedings{loshchilov2017,
 title={{SGDR}: Stochastic Gradient Descent with Warm Restarts},
 author={Loshchilov, Ilya and Hutter, Frank},
 year={2017},
 booktitle={International Conference on Learning Representations},
 url={https://arxiv.org/abs/1608.03983}
}

@inproceedings{cubuk2020,
 title={{RandAugment}: Practical Automated Data Augmentation with a Reduced Search Space},
 author={Cubuk, Ekin D. and Zoph, Barret and Shlens, Jonathon and Le, Quoc V.},
 year={2020},
 booktitle={Proceedings of the IEEE/CVF Conference on Computer Vision and Pattern Recognition Workshops},
 url={https://openaccess.thecvf.com/content_CVPRW_2020/html/w40/Cubuk_Randaugment_Practical_Automated_Data_Augmentation_With_a_Reduced_Search_Space_CVPRW_2020_paper.html}
}

@inproceedings{zhang2018,
 title={mixup: Beyond Empirical Risk Minimization},
 author={Zhang, Hongyi and Cisse, Moustapha and Dauphin, Yann N. and Lopez-Paz, David},
 year={2018},
 booktitle={International Conference on Learning Representations},
 url={https://openreview.net/forum?id=r1Ddp1-Rb}
}

@inproceedings{yun2019,
 title={{CutMix}: Regularization Strategy to Train Strong Classifiers with Localizable Features},
 author={Yun, Sangdoo and Han, Dongyoon and Oh, Seong Joon and Chun, Sanghyuk and Choe, Junsuk and Yoo, Youngjoon},
 year={2019},
 booktitle={Proceedings of the IEEE/CVF International Conference on Computer Vision},
 pages={6023--6032},
 url={https://openaccess.thecvf.com/content_ICCV_2019/html/Yun_CutMix_Regularization_Strategy_to_Train_Strong_Classifiers_With_Localizable_Features_ICCV_2019_paper.html}
}

@inproceedings{zhong2020,
 title={Random Erasing Data Augmentation},
 author={Zhong, Zhun and Zheng, Liang and Kang, Guoliang and Li, Shaozi and Yang, Yi},
 year={2020},
 booktitle={Proceedings of the AAAI Conference on Artificial Intelligence},
 volume={34},
 number={7},
 pages={13001--13008},
 url={https://ojs.aaai.org/index.php/AAAI/article/view/7000}
}

@inproceedings{huang2016,
 title={Deep Networks with Stochastic Depth},
 author={Huang, Gao and Sun, Yu and Liu, Zhuang and Sedra, Daniel and Weinberger, Kilian Q.},
 year={2016},
 booktitle={Proceedings of the European Conference on Computer Vision},
 url={https://arxiv.org/abs/1603.09382}
}

@inproceedings{szegedy2016,
 title={Rethinking the Inception Architecture for Computer Vision},
 author={Szegedy, Christian and Vanhoucke, Vincent and Ioffe, Sergey and Shlens, Jon and Wojna, Zbigniew},
 year={2016},
 booktitle={Proceedings of the IEEE Conference on Computer Vision and Pattern Recognition},
 pages={2818--2826},
 url={https://openaccess.thecvf.com/content_cvpr_2016/html/Szegedy_Rethinking_the_Inception_CVPR_2016_paper.html}
}

@inproceedings{ioffe2015,
 title={Batch Normalization: Accelerating Deep Network Training by Reducing Internal Covariate Shift},
 author={Ioffe, Sergey and Szegedy, Christian},
 year={2015},
 booktitle={Proceedings of the 32nd International Conference on Machine Learning},
 volume={37},
 pages={448--456},
 url={https://proceedings.mlr.press/v37/ioffe15.html}
}

@inproceedings{lakshminarayanan2017,
 title={Simple and Scalable Predictive Uncertainty Estimation using Deep Ensembles},
 author={Lakshminarayanan, Balaji and Pritzel, Alexander and Blundell, Charles},
 year={2017},
 booktitle={Advances in Neural Information Processing Systems},
 volume={30},
 url={https://proceedings.neurips.cc/paper_files/paper/2017/hash/9ef2ed4b7fd2c810847ffa5fa85bce38-Abstract.html}
}

@article{fort2019,
 title={Deep Ensembles: A Loss Landscape Perspective},
 author={Fort, Stanislav and Hu, Huiyi and Lakshminarayanan, Balaji},
 year={2019},
 journal={arXiv preprint arXiv:1912.02757},
 url={https://arxiv.org/abs/1912.02757}
}

@inproceedings{azulay2021,
 title={On the Implicit Bias of Initialization Shape: Beyond Infinitesimal Mirror Descent},
 author={Azulay, Shahar and Moroshko, Edward and Nacson, Mor Shpigel and Woodworth, Blake E. and Srebro, Nathan and Globerson, Amir and Soudry, Daniel},
 year={2021},
 booktitle={Proceedings of the 38th International Conference on Machine Learning},
 volume={139},
 pages={468--477},
 url={https://proceedings.mlr.press/v139/azulay21a.html}
}

@inproceedings{gissin2020,
 title={The Implicit Bias of Depth: How Incremental Learning Drives Generalization},
 author={Gissin, Daniel and Shalev-Shwartz, Shai and Daniely, Amit},
 year={2020},
 booktitle={International Conference on Learning Representations},
 url={https://openreview.net/forum?id=H1lj0nNFwB}
}

@article{jacot2022,
 title={Saddle-to-Saddle Dynamics in Deep Linear Networks: Small Initialization Training, Symmetry, and Sparsity},
 author={Jacot, Arthur and Ged, Fran{\c{c}}ois and {\c{S}}im{\c{s}}ek, Berfin and Hongler, Cl{\'e}ment and Gabriel, Franck},
 year={2022},
 journal={arXiv preprint arXiv:2106.15933},
 url={https://arxiv.org/abs/2106.15933}
}

@inproceedings{bantzis2026,
 title={Saddle-to-Saddle Dynamics in Deep {ReLU} Networks: Low-Rank Bias in the First Saddle Escape},
 author={Bantzis, Ioannis and Simon, James B. and Jacot, Arthur},
 year={2026},
 booktitle={International Conference on Learning Representations},
 url={https://openreview.net/forum?id=B4zcoLvjw0}
}
\endgroup
\clearpage
\onecolumn
\appendix
\begin{center}
{\Large\bfseries \thetitle\par}
\medskip
{\large Supplementary Material}
\end{center}
This supplement provides the protocols, evidence, and proofs supporting the main paper. \Cref{sec:visionprotocols} specifies the vision protocols. \Cref{sec:visionsupp} contains initialization-seed tests, complete ViT and ImageNet trajectories, and matched sample histories. \Cref{sec:cnn} gives the local KL statement and its GroupNorm ResNet-18 test. \Cref{sec:twolayerdetails,sec:twoproofs} develop and validate the two-layer phase mechanism. \Cref{sec:bnablation} collects the normalization and initialization ablations.

\section{Vision experimental protocols}\label{sec:visionprotocols}
This appendix specifies the paired training and evaluation procedures used in Sections~\ref{sec:setting} and~\ref{sec:analysis}. We first define the metrics and shared randomness, then give the architecture-specific recipes and numerical verification. All image experiments use five epochs of warmup followed by cosine decay \cite{loshchilov2017}. The theoretical experiments use their separately specified GF, GD, or SGD dynamics.

\subsection{Controlled comparisons and evaluation metrics}\label{sec:definitions}
Let $\mathcal D=\{(x_i,\widetilde y_i)\}_{i=1}^N$ be a fixed observed training set, with clean labels $y_i$ available for experimental analysis. Labels are corrupted once and remain fixed. Two replicas have the same architecture and optimizer configuration, share the observed labels, minibatch sequence and transformed inputs, and differ in initialization. Denote logits by $z_{jt}(x)$ and probabilities by $p_{jt}(x)=\operatorname{softmax}(z_{jt}(x))$, $j\in\{0,1\}$. The stochastic coupling is part of the experiment: conditional sensitivity to initialization is different from the total variability of fully independent training pipelines.

For an input distribution $Q$, define
\begin{align}
S_Q(t)&=\frac12\E_{X\sim Q}\!\left[\begin{aligned}
 &\KL(p_{0t}(X)\Vert p_{1t}(X))\\[-2pt]
 &{}+\KL(p_{1t}(X)\Vert p_{0t}(X))
\end{aligned}\right].\label{eq:skl}
\end{align}
An exact identity is
\begin{equation}
S(p_0,p_1)=\tfrac12(p_0-p_1)^T(z_0-z_1).\label{eq:klidentity}
\end{equation}
Mutual cross-entropy $H(p,q)=H(p)+\KL(p\Vert q)$ combines entropy and directional discrepancy, so it can decrease while symmetric KL increases.

We distinguish $Q=\widehat Q_N$, the uniform distribution on training inputs, from clean held-out inputs and the population distribution in the theory. Population cross-entropy and classification error evaluate predictions against the true labels. Unless specified otherwise, reported risks average the two replicas. The theoretical constructions specify their population distribution explicitly.

\subsection{CIFAR-10 architectures and training}
\subsubsection{Primary ResNet-18 recipe}\label{sec:primarysupp}
The main-text CIFAR-10 ResNet-18 experiments use GroupNorm \cite{wu2018} and Xavier-uniform classifier weights \cite{glorot2010}. The same recipe is used for the label-ratio comparison, the initialization ablation, and the instrumented sample analysis.

The CIFAR-10 ResNet-18 stem has kernel $3$, stride $1$, padding $1$, and no initial max-pool. GroupNorm has 32 groups, $\epsilon=10^{-5}$, and learned affine parameters initialized to one and zero. All convolutions use Kaiming-normal fan-out/ReLU initialization \cite{he2015}. The final classifier weight uses Xavier-uniform gain one. Its original default bias is retained. AdamW \cite{loshchilov2019} uses LR $0.001$, betas $0.9/0.999$, $\epsilon=10^{-8}$, and WD $0.01$ on all parameters. Batch size 128 gives 391 updates per epoch, including the smaller final batch. The budget is 200 epochs, or 78,200 updates. Linear warmup spans 1,955 updates, followed by cosine decay to zero. Augmentation is random crop with padding four, horizontal flip, and CIFAR normalization. Replicas consume the same transformed tensor, labels, and order. Model seeds are 123/124 for the screen and 223/224 for the confirmation pair. Label and data seeds are 3407 and 9001. The data loader uses persistent workers.

\subsubsection{ViT-Tiny/4}
The CIFAR-10 architecture comparison uses ViT-Tiny \cite{dosovitskiy2021} with $4\times4$ patches on $32\times32$ images: 12 blocks, dimension 192, three attention heads, and MLP dimension 768. We use the default timm initialization without pretrained weights. A DeiT-inspired recipe \cite{touvron2021} uses AdamW, batch size 256, 196 updates per epoch, and 300 epochs. The learning rate rises from $10^{-6}$ to $2.5\cdot10^{-4}$ over five epochs, then follows cosine decay to $10^{-5}$. Weight decay is $0.05$, excluding biases, one-dimensional parameters, and parameters exempted by the model. Betas and epsilon match the ResNet AdamW settings. DropPath uses stochastic depth \cite{huang2016} with rate $0.1$ in both variants. The hard-label variant uses crop and flip. The stronger variant adds RandAugment \cite{cubuk2020}, color jitter $0.4$, random erasing $0.25$ \cite{zhong2020}, Mixup $0.8$ \cite{zhang2018}, CutMix $1$ \cite{yun2019}, and label smoothing $0.1$ \cite{szegedy2016}. CIFAR padding crops replace ImageNet random-resized crops. This adapted recipe uses neither EMA nor distillation, and samples each training example once per epoch. Training uses BF16 autocast with FP32 parameters. Evaluation uses FP32 logits. The replicas share transformed inputs, mixed targets, and replayed forward RNG, including DropPath. They differ only in initialization.

\subsection{ImageNet-100 data and training}\label{sec:imagenetprotocol}
The public ImageNet-100 snapshot is \texttt{Yunzhe/ImageNet-100}, pinned to revision
\begin{quote}\small\texttt{e6653f7fb465c914166cf2edc06de43f384f4051}\end{quote}
Its 100 synsets match the CMC class list \cite{tian2020}, drawn from ImageNet \cite{deng2009}. Original encoded image bytes are retained without resizing during preparation. The fixed training set has 126,689 images and the validation set has 5,000, with 50 validation images per class. The data preparation inventory records source shard hashes and an ordered image manifest.

Both replicas train on identical transformed tensors and observed targets, with initialization seeds 123/124, data seed 9001, and target-construction seed 3407. Batch size is 256 and \texttt{drop\_last=True}, giving 494 updates per epoch. The 225 discarded tail examples change with each shuffle. Evaluation covers every image. Both recipes use five epochs of linear warmup from $10^{-6}$, followed by cosine decay. All weights are learned without pretraining.

ResNet-18 uses torchvision's default initialization and standard ImageNet stem, 200 epochs, SGD with momentum $0.9$, peak LR $0.1$, final LR zero, and weight decay $10^{-4}$ on all parameters. Training uses random-resized crops and horizontal flips. DeiT-Tiny/16 uses timm's default initialization, 300 epochs, AdamW with peak LR $2.5\cdot10^{-4}$, final LR $10^{-5}$, weight decay $0.05$ with bias/one-dimensional/model exemptions, and DropPath $0.1$. Its training recipe uses random-resized crops, flips, RandAugment, color jitter $0.3$, random erasing $0.25$, Mixup $0.8$, CutMix $1$, and label smoothing $0.1$, without EMA, distillation, or repeated-augmentation sampling. These reference recipes specify the architecture-dependent training conditions used in the comparison.

Training uses BF16 autocast with FP32 parameters. Evaluation uses FP32 forward logits and FP64 log-softmax metric arithmetic. Fixed evaluation views resize the short side to 256 and center-crop to 224, with bilinear interpolation for ResNet and bicubic for DeiT.

\subsection{Numerical verification}\label{sec:visionaudit}
\paragraph{CIFAR-10.}
The retained GroupNorm ResNet-18 experiments on CIFAR-10 use 201 integer-epoch observations and 209 total evaluations over all 50,000 training and 10,000 test images. Native training and forward logits use FP32. Metric probabilities and aggregation use FP64. Per-sample KL means reproduce the training logs within $2.3\cdot10^{-16}$. Independent final-checkpoint replay passes a continuous-metric tolerance of $5\cdot10^{-5}$ and a count tolerance of two examples. The additional initialization pairs in \cref{sec:seedablation} pass the same checks. Label hashes, shared input tensors, and initialization settings are audited. The matched sample-analysis run additionally passes the exact history replay in \cref{sec:matchedprimary}.

\paragraph{ImageNet-100.}
All integer epochs and metric finite values were checked. At initialization, the selected valley, peak, and endpoint, saved scalar KL over all 126,689 training images reproduces the corresponding logged mean within $1.1\cdot10^{-10}$. ImageNet verification consists of complete epoch coverage and saved-scalar aggregation at these four selected checkpoints. The CIFAR audits additionally include independent final-checkpoint-to-logit replay.

\FloatBarrier
\section{Additional vision evidence}\label{sec:visionsupp}
We supply the evidence supporting the vision results and sample analysis in Sections~\ref{sec:setting} and~\ref{sec:analysis}. We first vary initialization seeds, then give the complete ViT and ImageNet-100 trajectories, and finally detail the matched sample histories. The training protocols are in \cref{sec:visionprotocols}.

\subsection{Initialization-seed ablation}
\label{sec:seedablation}
\begin{figure}[htbp]\centering
\includegraphics[width=\linewidth]{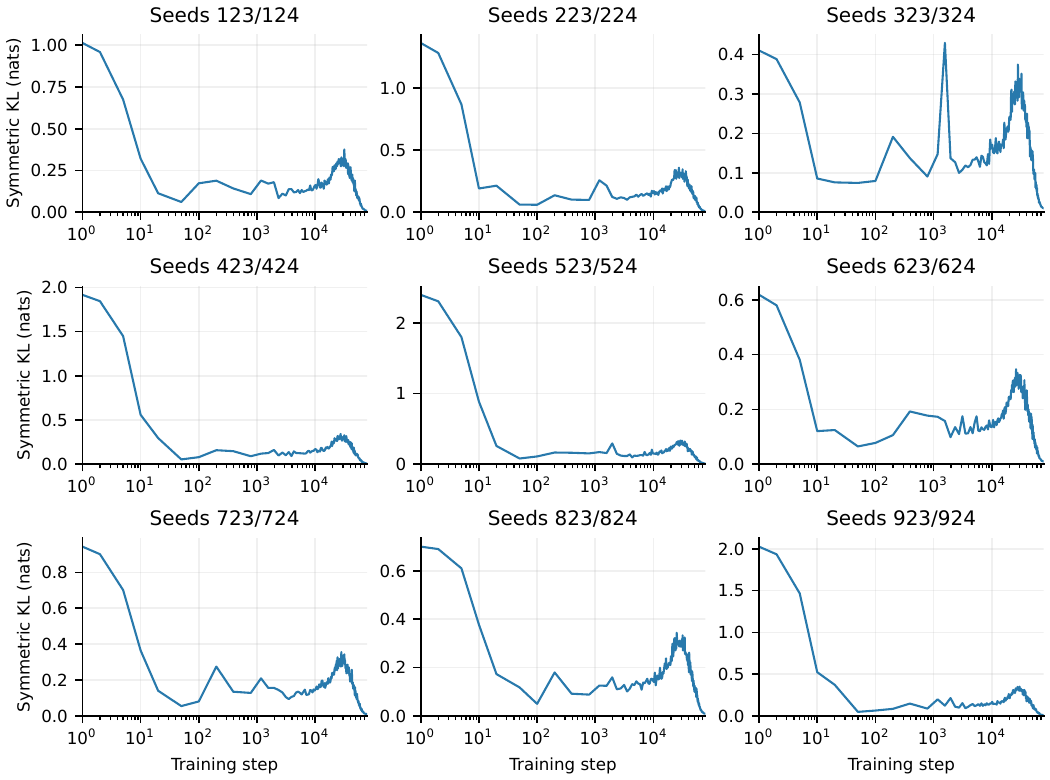}
\caption{CIFAR-10 ResNet-18 with 10\% label replacement across nine initialization pairs. The first two panels show the original and confirmation pairs, and the remaining seven show the additional prespecified pairs. All complete native trajectories are shown without smoothing, starting at step 1 on a log axis.}\label{fig:seedablation}
\end{figure}

We vary only initialization in the CIFAR-10 ResNet-18 experiment with 10\% label replacement, holding the training recipe, labels, minibatches, and augmented inputs fixed. Seven prespecified pairs supplement the original pair (123/124) and confirmation pair (223/224). All seven additional pairs exhibit an early decline, an intermediate resurgence, and a late decline. Their peak-to-valley ratios range from 3.40 to 3.78. Across all nine pairs, the early valleys occur between epochs 5 and 11 and the subsequent peaks between epochs 64 and 81, followed by a decline through epoch 200. \Cref{fig:seedablation,tab:seedablation} report every pair. All seven new runs complete the full budget and pass the independent final-checkpoint replay and all-sample aggregation checks in \cref{sec:visionaudit}.

\begin{table}[htbp]\centering\small
\caption{Initialization-seed ablation for ResNet-18 trained on CIFAR-10 with 10\% label replacement. Valleys are minima over integer epochs 5--30, and peaks are subsequent maxima through epoch 120. Final values are measured at epoch 200. All KL values are in nats.}\label{tab:seedablation}
\begin{tabular}{lrrrrr}\toprule
Seeds & Initial & Valley (epoch) & Peak (epoch) & Final & Peak/valley\\\midrule
123/124 & 1.0418 & 0.0848 (6) & 0.3761 (81) & 0.00952 & 4.44\\
223/224 & 1.3999 & 0.1001 (9) & 0.3585 (70) & 0.00967 & 3.58\\
323/324 & 0.4202 & 0.0998 (7) & 0.3746 (72) & 0.01018 & 3.75\\
423/424 & 1.9519 & 0.0909 (10) & 0.3414 (70) & 0.00657 & 3.76\\
523/524 & 2.4494 & 0.0954 (11) & 0.3358 (74) & 0.00805 & 3.52\\
623/624 & 0.6400 & 0.0985 (5) & 0.3469 (67) & 0.00957 & 3.52\\
723/724 & 0.9634 & 0.0942 (9) & 0.3558 (72) & 0.01019 & 3.78\\
823/824 & 0.7068 & 0.1013 (9) & 0.3443 (64) & 0.00923 & 3.40\\
923/924 & 2.0735 & 0.0927 (9) & 0.3508 (77) & 0.00914 & 3.78\\
\bottomrule\end{tabular}
\end{table}

\FloatBarrier
\subsection{ViT training recipes and label sources}\label{sec:vit}
\begin{figure*}[t]\centering
\includegraphics[width=\linewidth]{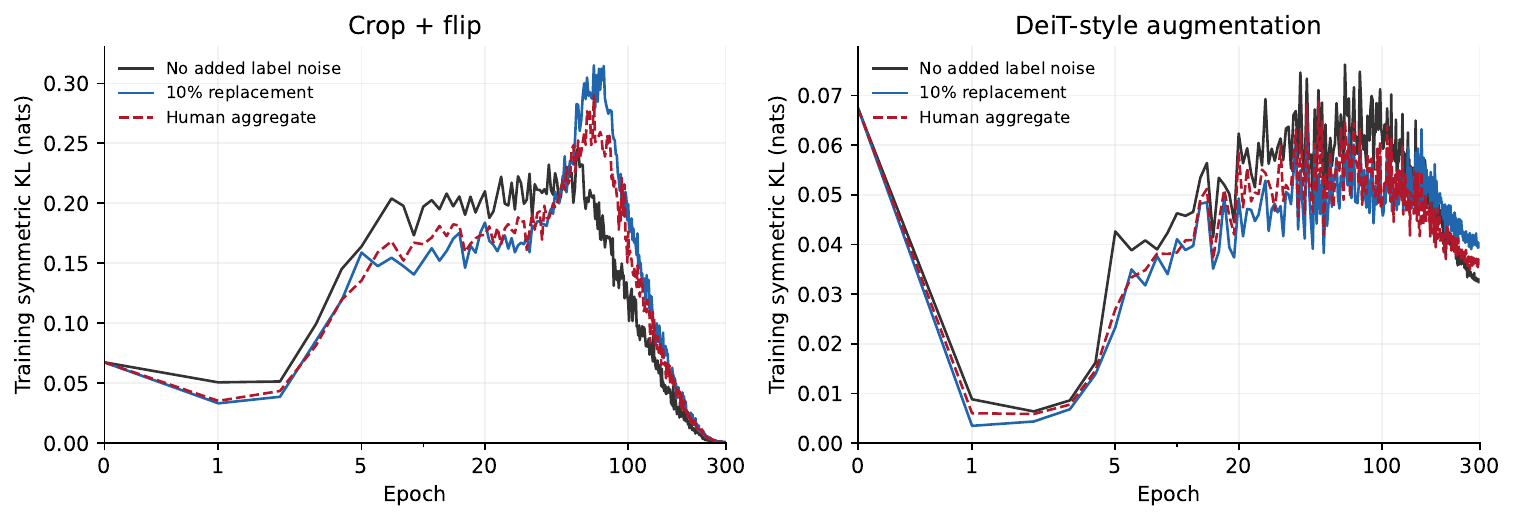}
\caption{ViT-Tiny/4 training symmetric KL on all 50,000 CIFAR-10 inputs, without smoothing. Each panel compares 10\% label replacement, CIFAR-10N aggregate annotations, and benchmark labels without injected noise. The epoch axis is linear from zero to one and logarithmic thereafter. One coupled initialization pair per condition. All runs complete 300 epochs and final checkpoints have been independently replayed.}\label{fig:vit}
\end{figure*}

Vision transformers also exhibit the trajectory. We train ViT-Tiny/4 on CIFAR-10 for 300 epochs with nominal $10\%$ synthetic replacement, CIFAR-10N aggregate annotations \cite{wei2022}, or no added label noise. The two noisy sources have actual wrong-label fractions $8.750\%$ and $9.010\%$. Both CIFAR-10 ViT-Tiny/4 recipes, using hard labels or stronger regularization, show an initial decrease, an intermediate resurgence, and a later decline (\cref{fig:vit}), including their zero-injection controls. The hard-label runs show a more pronounced resurgence and a stronger late decline. Strong regularization reduces their peaks and leaves larger nonzero tails; \cref{tab:vit} reports the corresponding trajectory summaries. The resurgence persists under both recipes.

\begin{table*}[t]\centering\small
\caption{ViT-Tiny/4 training KL on CIFAR-10 in nats. Valley is the minimum in integer epochs 1--50. Peak is the subsequent maximum through epoch 300. These selections are retrospective.}\label{tab:vit}
\begin{tabular}{llrrr}\toprule
Recipe&Labels&Valley (epoch)&Peak (epoch)&Final KL\\\midrule
Hard&No added noise&0.050643 (1)&0.246356 (57)&0.000223\\
Hard&Synthetic&0.033168 (1)&0.314970 (68)&0.000717\\
Hard&Human&0.035191 (1)&0.290030 (68)&0.000712\\
Strong&No added noise&0.006376 (2)&0.076231 (66)&0.032577\\
Strong&Synthetic&0.003474 (1)&0.066178 (43)&0.039506\\
Strong&Human&0.005859 (2)&0.068851 (49)&0.035933\\\bottomrule
\end{tabular}
\end{table*}
The stronger recipe optimizes mixed targets with label smoothing, which also define its late training objective. For ViT-Tiny/4 on CIFAR-10, mean final test accuracy is $89.46\%$ with synthetic labels and $89.96\%$ with human labels, versus $78.26\%$ and $78.83\%$ for the hard-label recipe. These accuracy comparisons hold the ViT architecture fixed and vary its training recipe. Complete epoch coverage and all-sample KL aggregation were checked, and final train/test KL, CE, and accuracy were independently recomputed from the saved checkpoints.

For ViT-Tiny/4 on CIFAR-10, label replacement raises the hard-label peak but lowers the strongly regularized peak (\cref{fig:vit,tab:vit}). In the latter comparison, peak-minus-valley also decreases, while peak/valley increases because the early valley is lower. Benchmark labels already give a clear resurgence, and the amplitude response to replacement depends on both the recipe and the statistic.

\subsection{ImageNet-100 trajectories}\label{sec:imagenet}
We extend the measurements to ImageNet-100, using the fixed CMC class list, 126,689 training images, 5,000 validation images, and $224\times224$ inputs. ResNet-18 and DeiT-Tiny/16 are trained from scratch with fixed paired initialization seeds. The complete recipes and evaluation views are specified in \cref{sec:imagenetprotocol}.

\begin{figure*}[t]\centering
\includegraphics[width=\linewidth]{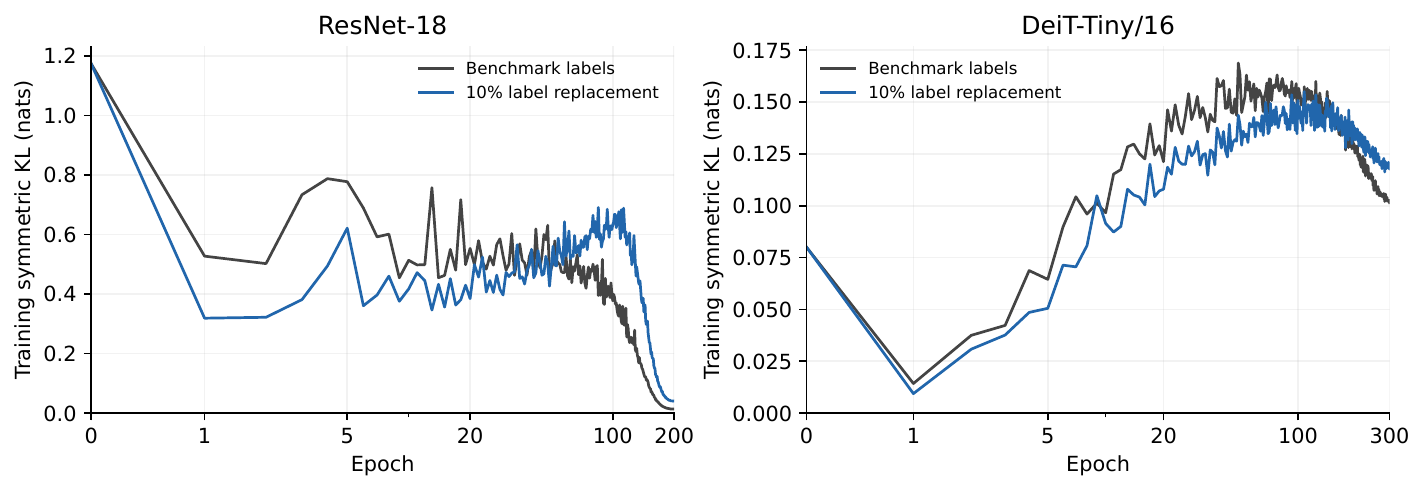}
\caption{Completed ImageNet-100 trajectories on fixed evaluation views of all training images. Each panel compares the original benchmark targets and $10\%$ uniform label replacement under one architecture's recipe. The epoch axis is linear from zero to one and logarithmic thereafter. Curves are unsmoothed. ResNet-18 runs complete 200 epochs and DeiT-Tiny/16 runs complete 300 epochs. Each condition uses one coupled initialization pair.}\label{fig:imagenet}
\end{figure*}

\Cref{fig:imagenet} shows an initial decline, an intermediate resurgence, and a later decline for both architectures and label conditions. DeiT-Tiny/16 retains greater discrepancy at the end of training. The detailed trajectory summaries in \cref{tab:imagenet} quantify the architecture-dependent amplitudes and peak times. The strongly regularized ImageNet-100 DeiT-Tiny/16 recipe optimizes mixed targets with label smoothing throughout training.

\begin{table}[tbp]\centering\small
\caption{Symmetric KL for ResNet-18 and DeiT-Tiny/16 on ImageNet-100 with benchmark labels and 10\% label replacement. Valley is selected retrospectively in integer epochs 1--50 and peak is the subsequent maximum.}\label{tab:imagenet}
\begin{tabular}{llrrrr}\toprule
Architecture&Targets&Initial&Valley (epoch)&Peak (epoch)&Final\\\midrule
ResNet-18&Benchmark&1.173413&0.454480 (9)&0.757367 (13)&0.013387\\
ResNet-18&10\% replacement&1.173413&0.318797 (1)&0.690693 (85)&0.041052\\
DeiT-Tiny/16&Benchmark&0.079956&0.014266 (1)&0.168796 (49)&0.101491\\
DeiT-Tiny/16&10\% replacement&0.079956&0.009392 (1)&0.155030 (108)&0.118668\\\bottomrule
\end{tabular}
\end{table}

\FloatBarrier
\subsection{Matched sample-level diagnostics}\label{sec:samples}\label{sec:matchedprimary}
This subsection specifies the histories, aggregation, and selection rules behind Figures~\ref{fig:mainsamples} and~\ref{fig:sampleforgetting}. All observations come from the GroupNorm ResNet-18 experiment on CIFAR-10 with 10\% label replacement used throughout Section~\ref{sec:analysis}.

To relate the aggregate trajectory, individual forgetting, and local response within one training condition, we rerun the fixed GroupNorm32/Xavier CIFAR-10 recipe in \cref{sec:primarysupp} with nominal 10\% label replacement and seeds 123/124. Section~\ref{sec:analysis} uses this instrumented run throughout. The Section~\ref{sec:setting} trajectory comparisons retain their original runs. The added records use logits already computed during evaluation and introduce no random draws or optimizer changes. Native cuDNN benchmarking permits different numerical trajectories between reruns. The 200-epoch run completes 78,200 updates with the same five-epoch warmup, labels, shared augmented tensors, AdamW settings, and FP32 training. The dataset contains 45,625 images with unchanged labels and 4,375 with changed labels.

\paragraph{Full-population histories and denominators.}
We retain symmetric KL, both observed-label probabilities, and both target logit margins for all 50,000 training images at every integer epoch from 0 to 200. Probabilities and KL use FP64 arithmetic on native logits. Positive target margin denotes a correct observed-label prediction. No recorded margin is exactly zero. Every saved KL mean reproduces its training log, and the two mean negative log target probabilities reproduce the two training cross-entropies. Independent checkpoint replay at epochs 20 and 200, matching CPU image normalization and paired evaluation order, reproduces all saved target probabilities, margins, and KL values exactly.

We choose the discrepancy peak as the largest full-training mean KL among integer epochs at least 10. It occurs at epoch 73, with mean KL $0.345659$. A prior confident fit requires observed-label probability at least $0.9$ for three consecutive integer epochs before the peak, excluding epoch 0. The clean one-correct group contains 1,936 images. Its summed symmetric KL is $6,322.8860$, the denominator for both main-text percentages. Of this mass, $85.12\%$ comes from samples whose currently incorrect replica had such a prior fit. For $50.73\%$, both replicas previously sustained this confidence together and the currently correct replica remained correct thereafter. Both percentages weight images by their KL within this group.

\paragraph{Individual trajectory selection.}
To illustrate sustained asymmetric forgetting, we consider clean images with exactly one correct replica at epoch 73, a prior joint three-epoch confident fit, a retained replica that remains correct after that fit, and a run of at least three consecutive incorrect predictions containing epoch 73 for the other replica. Among the 61 eligible images, we select the lower median by epoch-73 KL, breaking ties by image index. This deterministic rule selects image 29006. Both replicas meet the confidence condition at epochs 63--65. Seed 124 is incorrect at epochs 71--73 while seed 123 remains correct. The replicas agree again at the end of training. Figure~\ref{fig:sampleforgetting} shows all 201 raw integer-epoch observations, without smoothing. Figure~\ref{fig:mainsamples} instead smooths each sample's KL over five centered integer epochs, truncated at the endpoints, before differencing and independently sorting each interval into 256 displayed percentiles. The aggregation identity has maximum numerical error $8.4\cdot10^{-17}$.

\clearpage
\section{Local prediction geometry and CNN controls}\label{sec:cnn}
We give the assumptions and proof for the local KL expansion used in Section~\ref{sec:analysis}, together with the matched GroupNorm ResNet-18 intervention. The comparison tests how function-equivalent predictors respond differently to the same training examples.

\subsection{Regenerating KL from an equal-function state}
Consider two models whose current query probabilities agree, so that $p_0(x)=p_1(x)=p(x)$. The models may use different parameterizations. Assume logits are three times continuously differentiable locally, with bounded derivatives and sample gradients on the finite inputs under consideration. Let
\begin{equation}
J_j(x)=\partial_{\theta_j}z_j(x),\quad F_x=\diag(p(x))-p(x)p(x)^T,\quad
h_i(x)=J_0(x)g_{0i}-J_1(x)g_{1i},
\end{equation}
where $g_{ji}=\nabla_{\theta_j}\ell_i$. Take a shared iid batch of size $B$, and ordinary SGD $\theta_j^+=\theta_j-\eta g_j(B)$.
\begin{proposition}[Local KL regeneration]\label{prop:local}
Under these assumptions,
\begin{equation}
\E_B S_Q^+=\frac{\eta^2}{2}\E_{x\sim Q}\left[
\bar h_x^TF_x\bar h_x+\frac1B\tr\big(F_x\Cov_i h_i(x)\big)\right]+O(\eta^3),\label{eq:local}
\end{equation}
where $\bar h_x=\E_i h_i(x)$. If predictions also agree on the training inputs, CE gradients give
\begin{equation}
h_i(x)=[K_0(x,x_i)-K_1(x,x_i)](p(x_i)-e_{\widetilde y_i}),\quad
K_j(x,x_i)=J_j(x)J_j(x_i)^T.
\end{equation}
\end{proposition}
The proof follows by expanding \cref{eq:klidentity} at equal probabilities and using the exact second moment of a batch average. Details are in \cref{sec:prooflocal}. The formula combines local KL/Fisher geometry with the SGD second-moment identity. The criterion uses each model's prediction response in the shared output space.

An increase requires the different kernel responses to the residual to survive the Fisher nullspace. The mean-update term remains for full-batch training, and the covariance contribution scales as $1/B$. 

For a ReLU network, this local expansion applies on a neighborhood with fixed activation masks. We test the response with ordinary SGD at saved checkpoints of the GroupNorm ResNet-18 training run.

\subsection{Matched GroupNorm ResNet-18 intervention}\label{sec:matchedresponse}
We test the local prediction at the checkpoints used for the sample analysis in \cref{sec:matchedprimary}, so that trajectory diagnostics and response interventions share the same GroupNorm ResNet-18 training condition on CIFAR-10 with 10\% label replacement.

To isolate update geometry at fixed predictions, we take replica 123 at epochs 20 and 200 of this same run. Multiplying every GroupNorm affine weight and bias by $c=2$, dividing every convolution weight except the stem by $c$, and dividing the classifier weight by $c$ preserves the function. Stem weights and classifier bias remain unchanged. Every normalization input remains unchanged, including in residual downsampling branches, so GroupNorm's $\epsilon$ is preserved. FP64 query logits agree exactly before the intervention. The $c=1$ identity control uses the same procedure without rescaling.

We draw 256 distinct held-out CIFAR-10 test queries and eight iid training batches of size 128 with replacement, using a fixed RNG seed 911931. All inputs are unaugmented and fixed. Both parameterizations receive the same batch at each trial. Each trial starts from the saved parameters and applies one ordinary Euclidean SGD step at $\eta\in\{10^{-5},3\cdot10^{-5},10^{-4},3\cdot10^{-4},10^{-3},3\cdot10^{-3},10^{-2}\}$. Training labels are the run's fixed replaced labels. Parameters remain immutable between interventions, verified by hashes. The intervention measures ordinary SGD responses at these saved checkpoints.

For each batch, automatic JVPs give the difference $d(x)$ of the two logit update directions. We compare actual symmetric KL with $\eta^2 C$, where $C=\frac12\E_Q[d^TF_xd]$. Table~\ref{tab:matchedresponse} reports means over the eight batches at $\eta=10^{-4}$. The relative error divides the absolute difference of those means by the mean quadratic prediction. The largest individual-batch relative errors are reported separately. Independent NumPy replay of all saved logits and JVPs agrees with the recorded KL within $4.4\cdot10^{-19}$. Both identity controls give exactly zero KL for every batch and step size. These finite-step checks quantify the mean and largest batchwise deviations in the eight sampled ordinary-SGD interventions.

\begin{table}[htbp]\centering\small
\caption{Matched GroupNorm ResNet-18 response controls at step size $10^{-4}$. Actual and predicted KL are means over eight shared batches and 256 fixed queries. Relative errors compare mean responses; the last column reports the largest batchwise error.}\label{tab:matchedresponse}
\begin{tabular}{rrrrr}\toprule
Epoch & Actual KL & Quadratic prediction & Mean error & Max batch error\\\midrule
20 & $1.2103680\cdot10^{-7}$ & $1.2103021\cdot10^{-7}$ & $0.0054\%$ & $0.0479\%$\\
200 & $3.6610353\cdot10^{-12}$ & $3.6609588\cdot10^{-12}$ & $0.0021\%$ & $1.0009\%$\\\bottomrule
\end{tabular}
\end{table}

\subsection{Proof of local KL regeneration}\label{sec:prooflocal}
At equal probabilities, logits can differ by an input-dependent all-ones shift. The shift drops out of \cref{eq:klidentity}. For a shared batch $\mathcal B$ of size $B$, write $h_B=B^{-1}\sum_{i\in\mathcal B}h_i$. An SGD step changes logits by $-\eta J_jg_j(B)+O(\eta^2)$ and the probability difference by $-\eta F_xh_B+O(\eta^2)$. Their product gives $\eta^2h_B^TF_xh_B/2+O(\eta^3)$. The iid batch second moment is
\[
\E h_Bh_B^T=\bar h\bar h^T+\Cov(h_i)/B,
\]
which proves \cref{prop:local}. Uniform finite-data local bounds control the remainder.

\FloatBarrier
\section{Two-layer model and numerical validation}\label{sec:twolayerdetails}
Early agreement can hide initialization-dependent information in a weakly expressed direction. We analyze fully trained two-layer linear networks in which strong-direction fitting initially aligns predictions and a remaining target correlation drives weak-direction learning. The learned response amplifies the weak coefficient, exposing different initialization-dependent phases on a common learning curve. Prediction discrepancy reflects these phase differences, while population risk depends on the target dependence acquired along the curve.

The scalar limits derived below have the common form
\begin{equation}
 \begin{aligned}
 M'&=2M\mathcal F(M),\\
 \mathcal F(m)&=-\partial_m\mathcal L(s^*(m),m).
 \end{aligned}
 \label{eq:sharedphaseoverview}
\end{equation}
where $s^*(m)$ is the conditional strong fit and $\mathcal F$ is the remaining empirical weak residual. The data determine this residual and the risk interpretation. The constructions below supply explicit assumptions under which strong tracking, amplification, and the initialization phases can be derived. \Cref{eq:sharedphaseoverview} summarizes the phase limits for these prototype tasks.

\subsection{Network, inputs, and empirical targets}
Consider the width-one two-layer linear network
\begin{equation}
 f_\theta(x)=u(v_1x_1+v_2x_2),\qquad \theta=(u,v_1,v_2).
 \label{eq:twomodel}
\end{equation}
All three weights are trained and initialized independently as $\mathcal N(0,1)$. The clean input distribution is uniform on $x=b(1,at)$, with independent $b,t\in\{-1,1\}$ and true label $b$. The amplitude $a>0$ makes the second direction initially weak. The training design has balanced weak groups, and a uniformly chosen fixed subset of $\rho N$ labels is flipped independently of the inputs. Write the observed label as $b_i\zeta_i$, and define
\begin{equation}
 \begin{aligned}
 &s=uv_1,\qquad m=auv_2,\\
 &T_i=(1+\zeta_i)/2,\quad q_\pm=\E_i[T_i\mid t_i=\pm1],\\
 &\bar q=(q_++q_-)/2=1-\rho,\\
 &c=(q_+-q_-)/2.
 \end{aligned}
 \label{eq:twocoef}
\end{equation}
Expectations indexed by $i$ refer to the fixed empirical sample. Multiplication by the true class normalizes the logit to $z_i=s+t_im$ and preserves binary CE and symmetric KL. The effective training loss is
\begin{equation}
 \mathcal L(s,m)=\frac12\sum_{r\in\{-1,1\}}
 \left[\log(1+e^{s+rm})-q_r(s+rm)\right].
 \label{eq:twoloss}
\end{equation}
We condition on $0<c<\rho<1/2$. Reversing $t$ covers negative $c$, and $c=0$ supplies a control. The correlation $c=N^{-1}\sum_i t_iT_i$ is a realized imbalance from input-independent corruption. Bounds that remain nonzero as $a\to0$ keep this finite dataset and $c$ fixed.

The function class is linear, but the factors learn nonlinearly. This distinction is central: a linear predictor with the same initial function provides a control for the effect of parameterization. Staged learning and initialization-dependent delays in factorized networks have established precedents \cite{saxe2014,pesme2023,even2023}. Our target is their implication for replica discrepancy and clean risk.

\subsection{Strong alignment and amplification of the weak response}
Let
\begin{equation}
 \begin{aligned}
 h(s,m)&=\tfrac12[\sigma(s+m)-\sigma(s-m)],\\
 A&=-\partial_s\mathcal L,\\
 B&=-\partial_m\mathcal L=c-h(s,m).
 \end{aligned}
\end{equation}
Full-batch gradient flow induces the exact effective dynamics
\begin{equation}
 \begin{gathered}
 \binom{\dot s}{\dot m}=K\binom{A}{B},\\
 K=\begin{pmatrix}
 u^2+v_1^2&av_1v_2\\
 av_1v_2&a^2(u^2+v_2^2)
 \end{pmatrix}.
 \end{gathered}
 \label{eq:twokernel}
\end{equation}
The output weight $u$ couples the two directions. Holding the strong coordinate stationary locally gives the weak response coefficient
\begin{equation}
 g=K_{mm}-\frac{K_{sm}^2}{K_{ss}}
   =a^2u^2\left(1+\frac{v_2^2}{u^2+v_1^2}\right).
 \label{eq:twogain}
\end{equation}
The off-diagonal terms retain the coupling through the shared output weight.

At $a=0$, strong fitting brings $s$ to $s_0=\operatorname{logit}(\bar q)$ for almost every initialization. For small $a$, it tracks the conditional strong optimum $s^*(m)$, defined by
\begin{equation}
 \sigma(s^*(m)+m)+\sigma(s^*(m)-m)=2\bar q.
 \label{eq:twoconditional}
\end{equation}
The weak coefficient is initially $O(a)$, so predictions first approach the common predictor $(s_0,0)$. The effective predictions align while the raw weights retain initialization-dependent information.

Gradient flow conserves
\begin{equation}
 D=u^2-v_1^2-v_2^2.
 \label{eq:twobalance}
\end{equation}
Consequently, with $U=u^2$,
\begin{equation}
 \begin{gathered}
 U=\frac{D+\sqrt{D^2+4s^2+4m^2/a^2}}2,\\
 v_1^2=s^2/U,\qquad v_2^2=m^2/(a^2U).
 \end{gathered}
 \label{eq:twoU}
\end{equation}
For bounded $D,s$ and fixed $m>0$, this gives $U\sim m/a$ and
\begin{equation}
 g=2am+o(a).
\end{equation}
The weak response thus grows from $O(a^2)$ at initialization to $O(a)$ after substantial weak fitting. Near $m=0$, its mean residual forcing is $B=c$, independent of the strong coefficient. A common empirical correlation therefore drives a learned gain whose activation time still depends on initialization.

\subsection{Initialization becomes a learning phase}
Write $H(m)=h(s^*(m),m)$ and $F_0(m)=c-H(m)$. The function $H$ increases from $0$ to $\rho$ on $m\ge0$. The unique positive root of $F_0$ and the corresponding strong optimum are
\begin{equation}
 \begin{aligned}
 m_* &=\tfrac12[\operatorname{logit}(q_+)-\operatorname{logit}(q_-)],\\
 s_* &=\tfrac12[\operatorname{logit}(q_+)+\operatorname{logit}(q_-)].
 \end{aligned}
 \label{eq:twooptimum}
\end{equation}
In slow time $\tau=at_{\rm GF}$, the amplified weak direction follows the limiting equation
\begin{equation}
 M'=2M[c-H(M)],\qquad
 \lim_{t\to-\infty}e^{-2ct}M(t)=1.
 \label{eq:twoouter}
\end{equation}
This normalized orbit increases from $0$ to $m_*$.

The phase on this orbit is determined before weak fitting. For raw initialization $\theta_0$, let $D_0$ be given by \cref{eq:twobalance}, and put
\begin{equation}
 \begin{aligned}
 w_0&=\operatorname{sign}(u_0+v_{10})v_{20},\\
 \Gamma_D(w)&=\left[\frac{D+w^2+\sqrt{(D+w^2)^2+4s_0^2}}2\right]^{1/2},\\
 \beta(\theta_0)&=\lim_{R\to\infty}R^2
 \exp\left[-2\int_{w_0}^R\frac{\dd w}{\Gamma_{D_0}(w)}\right],\\
 \chi(\theta_0)&=\frac{\log\beta(\theta_0)}{2c}.
 \end{aligned}
 \label{eq:twophasecoef}
\end{equation}
The sign fixes a function-preserving gauge. After strong alignment, the inner weak dynamics satisfy $\dd w/\dd\tau=c\Gamma_{D_0}(w)$. Since $\Gamma_D(w)\sim w$ for large positive $w$, they give $m\sim a\beta e^{2c\tau}$. Thus $\chi$ is computed directly from the raw initialization and the training dynamics.

The central phase statement is Theorem~\ref{thm:twophase} in the main paper.

The proof in \cref{sec:twoproofs} matches the fast strong fit, the initialization-dependent inner dynamics, and the outer orbit. A compact initialization event excluding the Gaussian-null unstable-mode boundary admits a constant $\nu>0$ for which full-batch GD has the same phase limit, with times $k=\lfloor(\tau_a+t)/(\eta a)\rfloor$, whenever $0<\eta\le\nu a$. This step restriction controls the strong response $K_{ss}=O(1/a)$ after weak fitting. The admissible step size therefore scales linearly with $a$.

\subsection{Discrepancy can vanish while clean risk grows}
The phase limit makes both observables explicit. On the conditional strong curve, the two group probabilities are $p_\pm(m)=\bar q\pm H(m)$. For two replicas at weak coefficients $m_0,m_1$, their balanced-group symmetric KL is exactly
\begin{equation}
 \mathcal S(m_0,m_1)
 =\tfrac12(m_0-m_1)[H(m_0)-H(m_1)].
 \label{eq:twoKLreduced}
\end{equation}
The strong-logit difference cancels because the two group probabilities have the same mean. Since $H$ is strictly increasing, different phases give positive KL at intermediate times. At the two ends of the common orbit, both replicas approach respectively $0$ and $m_*$, and \cref{eq:twoKLreduced} vanishes.

The three-stage comparison is Corollary~\ref{cor:twothree} in the main paper.

Independent initializations give unequal phases almost surely. Quantiles of this phase difference yield the nonvanishing intermediate gap. The timing differences emerge from the initialization-dependent inner dynamics on a common orbit.

Clean population CE on the \emph{same input distribution} has a different behavior. Along the conditional strong curve,
\begin{equation}
 \begin{gathered}
 \mathcal R(m)=-\tfrac12\log[\bar q^2-H(m)^2],\\
 \mathcal R'(m)>0\quad(m>0).
 \end{gathered}
 \label{eq:twoRiskreduced}
\end{equation}
Thus clean CE increases along each limiting weak-learning trajectory, even though replica discrepancy ultimately disappears. The exact early--late risk gap is
\begin{equation}
 \mathcal R(m_*)-\mathcal R(0)
 =-\tfrac12\log\left(1-\frac{c^2}{\bar q^2}\right)>0.
 \label{eq:tworiskgap}
\end{equation}
The fitted dependence on $t$ is spurious for the clean population but common to all seeds. Prediction sensitivity to initialization disappears on this population while the shared dependence remains. If $q_-<1/2<\bar q$, clean classification error also changes from $0$ to $1/2$. At lower noise, both classifiers can remain correct, separating CE from argmax error. These identities evaluate population risk on the four prototypes.

\subsection{A predictive weak direction}\label{sec:cleanphase}
The same response and phase dynamics can be instantiated with a weak direction carrying genuine task information. Keep the fully trained network and inputs $x=b(1,at)$, take $b$ uniform and $\PP(t=+1)=\pi\in(1/2,1)$, and set the \emph{observed and true label} to $Y=bt$ deterministically. A fixed training design matches these four population masses. The first coordinate is a predictive shortcut of accuracy $\pi$. The second is a smaller-scale, perfectly predictive feature. In this construction $b$ is the reference sign and $bt$ is the true class.

After sign normalization, the empirical loss is
\begin{equation}
 \begin{aligned}
 \mathcal L_{\rm cl}(s,m)={}&\pi\log(1+e^{-s-m})\\
 &{}+(1-\pi)\log(1+e^{s-m}).
 \end{aligned}\label{eq:cleanloss}
\end{equation}
Its conditional strong optimum $s^*_{\rm cl}(m)$ satisfies
\begin{equation}
 \begin{aligned}
 J(m)&:=\pi[1-\sigma(s^*_{\rm cl}(m)+m)]\\
 &=(1-\pi)\sigma(s^*_{\rm cl}(m)-m).
 \end{aligned}\label{eq:cleanflux}
\end{equation}
The positive function $J$ decreases with $m$. The residual weak forcing is $2J(m)$, with $b_0:=2J(0)=2\pi(1-\pi)>0$. The GF matching argument gives a normalized orbit
\begin{equation}
 \begin{gathered}
 M_{\rm cl}'=4M_{\rm cl}J(M_{\rm cl}),\\
 \lim_{t\to-\infty}e^{-2b_0t}M_{\rm cl}(t)=1.
 \end{gathered}\label{eq:cleanouter}
\end{equation}
The raw-initialization phase is obtained from \cref{eq:twophasecoef} with $s_0=\operatorname{logit}(\pi)$ and $c$ replaced by $b_0$. In shifted GF time, the replicas follow $M_{\rm cl}(t+\chi_j)$. The finite-window phase limit and its three-stage KL consequence are derived in \cref{sec:cleanproof}. This extension is stated for GF.

The same phases now have a different risk interpretation. On the conditional curve,
\begin{equation}
 \begin{aligned}
 \mathcal S_{\rm cl}(m_0,m_1)
   &=-(m_0-m_1)[J(m_0)-J(m_1)],\\
 \mathcal R_{\rm cl}'(m)&=-2J(m)<0.
 \end{aligned}\label{eq:cleanklrisk}
\end{equation}
KL tends to zero at both ends of the limiting phase window and is positive between unequal phases. Here the acquired weak dependence improves clean CE. The discrepancy geometry is the same: phase differences determine separation along the curve, while the curve's relation to the true task determines its risk.

\begin{figure*}[tp]\centering\includegraphics[width=\linewidth]{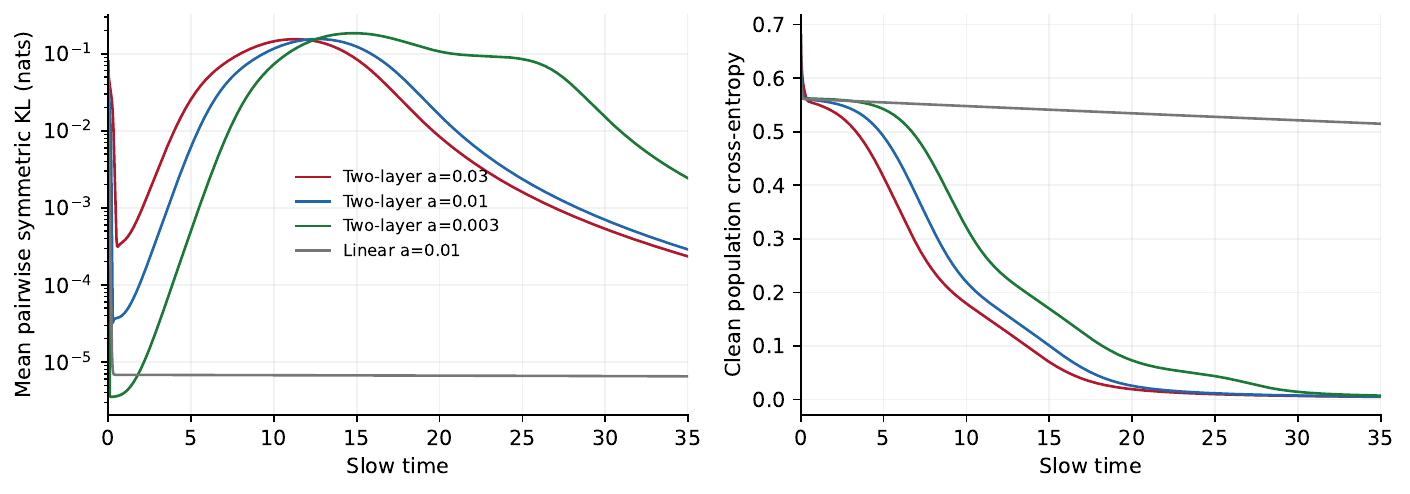}
\caption{Completely uncorrupted, separable two-layer task: KL resurgence with decreasing clean population CE. All three raw weights learn. The linear control matches their initial function. Curves average pairs among 16 models. Independent PyTorch two-layer updates match the numerical engine within $2.3\cdot10^{-16}$ over the first 512 updates of two models per nonlinear configuration.}\label{fig:cleanphasevalidation}
\end{figure*}

We check this case with full-batch GD at $\pi=0.75$, 16 independent unit-Gaussian models, fixed step $\eta=a/4$, and slow-time budget $a\eta k=35$. In \cref{fig:cleanphasevalidation}, the two-layer replicas first align, separate as the predictive weak direction is learned, and align again while population CE continues to improve. The function-matched single-layer control remains closely aligned over this interval. \Cref{tab:cleanphasetiming} tests the raw phase prediction of the $m=1$ crossing. Prediction is accurate at the smallest input scale and less accurate at the two larger scales, making the finite-scale dependence explicit. The 120 pair averages share models and are dependent. Small-step GD numerically tests the GF prediction.

\begin{table}[htbp]\centering\small
\caption{Raw phase predictions for the $m=1$ crossing in the task with a predictive weak direction. Each scale uses 16 Gaussian-initialized models, and all crossings are observed. MAE is in slow-time units.}\label{tab:cleanphasetiming}
\begin{tabular}{rrr}\toprule
Weak-input scale $a$ & Spearman correlation & Crossing-time MAE\\\midrule
$0.003$ & $1.000000$ & $0.008353$\\
$0.01$ & $0.647059$ & $1.267505$\\
$0.03$ & $0.644118$ & $1.297224$\\
\bottomrule
\end{tabular}
\end{table}

\subsection{Quantitative predictions and targeted controls}
For a threshold $m_{\rm cut}\in(0,m_*)$, the phase law predicts its first crossing time $T_{j,a}$:
\begin{equation}
 aT_{j,a}-\tau_a\longrightarrow G(m_{\rm cut})-\chi_j,
 \qquad G=M^{-1}.
 \label{eq:twohittime}
\end{equation}
For GD, replace $aT_{j,a}$ by $\eta aT_{j,a}$. Here fitting time is the first crossing of the specified learned-coefficient threshold.

For finite-$a$ prediction, we retain the full balance relation in \cref{eq:twoU}. Following $s=s^*(m)$ gives
\begin{equation}
 \frac{\dd m}{\dd\tau}=
 \frac{K_{ss}K_{mm}-K_{sm}^2}{a[K_{ss}-s^{*\prime}(m)K_{sm}]}[c-H(m)].
 \label{eq:tworeduction}
\end{equation}
The initial condition comes from the raw weights' fast strong fit, as detailed in \cref{sec:twoprediction}. The reduction, including its time origin and amplitude, is computed from the dataset and initialization.

\begin{figure*}[tp]\centering\includegraphics[width=\linewidth]{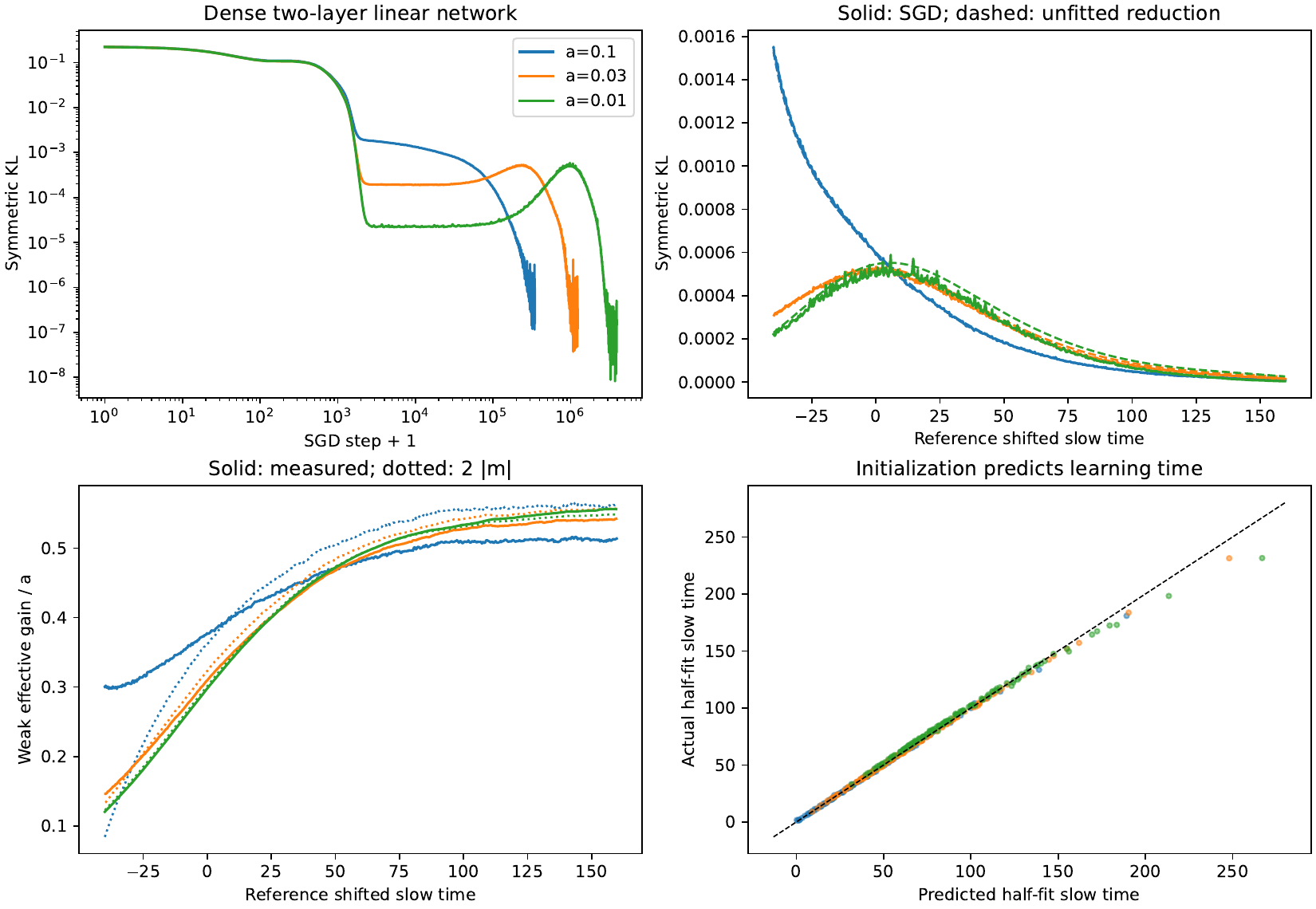}
\caption{Two-layer learning phases. Measured weak-response amplification and initialization-dependent half-fit times test the mechanism underlying the KL resurgence. Dashed curves use the reduction computed from raw initialization and data. The $a=0.1$ trajectory illustrates insufficient phase separation.}\label{fig:twomechanism}
\end{figure*}

\Cref{fig:twomechanism} tests the reduced dynamics in the fixed binary task with 10\% label flips. As the weak response grows, initialization-dependent learning times produce a renewed discrepancy peak. The reduction follows the intermediate rise and late decline for the two smaller input scales, while the largest scale shows overlapping stages. The predicted half-fit times closely track their measured values across initializations. Configurations, trajectory summaries, and prediction errors are collected in \cref{sec:twonumerics}.

Two controls remove separate components. Balancing flip counts sets $c=0$. Under gradient flow, $\dd(v_2^2)/\dd t=-2m h(s,m)\le0$, preventing the weak instability. A single-layer predictor starts at the \emph{same function}, with coefficients $u_0v_{10}$ and $u_0v_{20}$, but has constant weak gain $a^2$. Both controls have substantially smaller discrepancy than the two-layer peak. The function-matched comparison identifies an effect of learning parameterization within the same linear function class.

\begin{figure*}[tp]\centering\includegraphics[width=0.85\linewidth]{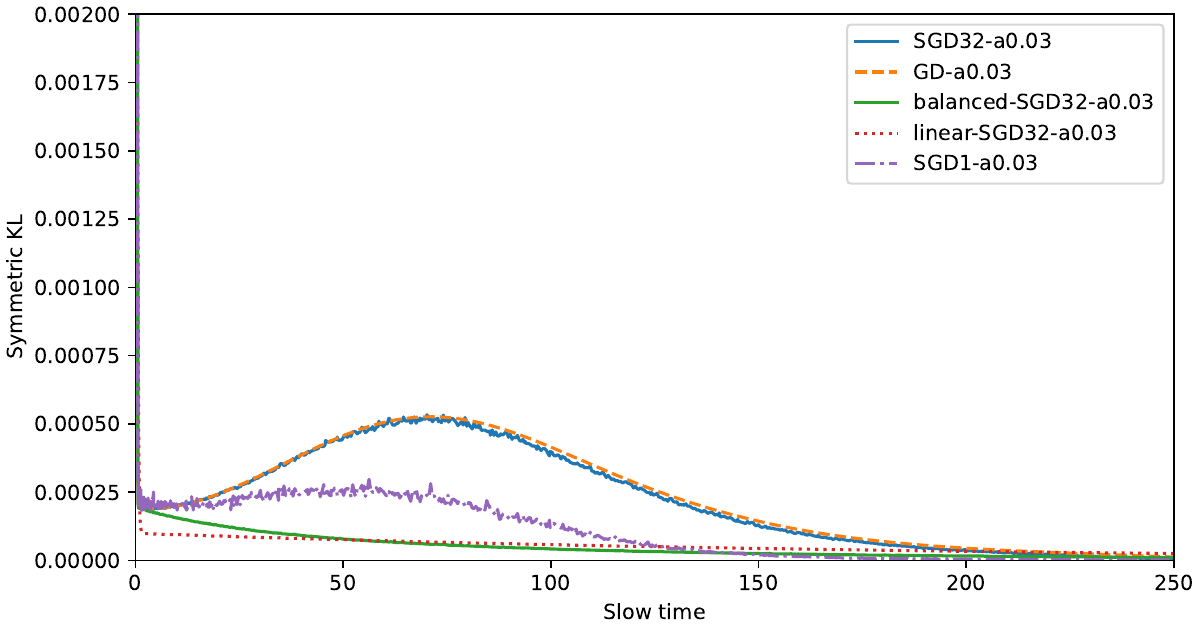}
\caption{Controls for empirical forcing, learned amplification, and stochastic corrections. Balanced corruption removes $c$. The function-matched linear predictor removes the changing weak gain. The batch-one run probes the effect of stronger sampling noise.}\label{fig:twocontrols}
\end{figure*}

\paragraph{Stochastic corrections.}
The phase proof covers gradient flow and stable full-batch GD. For a minibatch $\mathcal B$, let $A_{\mathcal B}=\operatorname{mean}_{i\in\mathcal B}r_i$ and $B_{\mathcal B}=\operatorname{mean}_{i\in\mathcal B}t_ir_i$. Actual SGD obeys the exact identity
\begin{equation}
 \begin{aligned}
 D^+-D=\eta^2\big[&(v_1A_{\mathcal B}+av_2B_{\mathcal B})^2\\
 &{}-u^2(A_{\mathcal B}^2+a^2B_{\mathcal B}^2)\big].
 \end{aligned}
 \label{eq:twostochasticbalance}
\end{equation}
Its conditional mean includes signed gradient-variance terms that correct the GF balance under SGD. At $a=0.03$, reducing batch size from $32$ to $1$ increases the median balance drift before half-fitting from $0.0866$ to $2.38$ and lowers the relative-time prediction correlation from $0.99945$ to $0.93078$. The SGD tests therefore measure both the prediction and its correction. The dense construction establishes a phase mechanism in a fully trained network with a preassigned weak direction. The CIFAR sample diagnostics additionally track individual fitting, confidence changes, and asymmetric forgetting.

\subsection{Computing the finite-amplitude prediction}\label{sec:twoprediction}
Following $s=s^*(m)$ in the effective flow imposes $\dot s=s^{*\prime}(m)\dot m$. Solving the two effective equations for $A$ gives the mobility in \cref{eq:tworeduction}. The conditional strong root can be computed directly: with $C=\cosh m$ and $d=(2\bar q-1)C$,
\[
 s^*(m)=\log\frac{d+\sqrt{d^2+4\bar q(1-\bar q)}}{2(1-\bar q)}.
\]
For raw initialization, put $C_0=u_0^2-v_{10}^2$. The $a=0$ strong fit gives
\[
 \widehat u_0=\operatorname{sign}(u_0+v_{10})
 \left[\frac{C_0+\sqrt{C_0^2+4s_0^2}}2\right]^{1/2}.
\]
The quantitative reduction starts at $m(0)=a\widehat u_0v_{20}$ and uses the raw value $D_0$. This idealizes the short initial strong transient. It is a finite-amplitude approximation to the learning dynamics, and retains no fitted trajectory parameters. In particular, treating $D_0$ as fixed is exact for GF, controlled for stable GD by \cref{eq:twoDGD}, and a testable approximation for SGD.

\subsection{Simulation protocol and numerical checks}\label{sec:twonumerics}
The main experiment uses $N=80$ inputs and label-flip seed 3407, which selects eight fixed flips independently of the pre-existing weak groups. Group counts are three and five, giving $c=0.025$. Four shared iid streams each train 32 independent unit-Gaussian models. Plain SGD uses batch size 32 and constant $\eta=0.01$. Full-batch GD uses the same learning rate. The same raw Gaussian draws and common iid stream prefixes are used across amplitudes and optimizers. There is no normalization, momentum, weight decay, warmup, schedule, augmentation or clipping. Statistics average distinct model pairs within each stream and then the four streams. Those pairs are dependent.

\begin{table}[tbp]\centering\small
\caption{Dense two-layer trajectories. Valley/peak statistics are descriptive saved-grid summaries. At $a=0.1$ the stages overlap and the measured SGD rise is negligible. Pair averages within a stream are dependent.}\label{tab:tworesults}
\begin{tabular}{lrrrr}\toprule
Setting&Initial KL&Valley&Later maximum&Endpoint\\\midrule
SGD 32, $a=0.1$&$0.224604$&$0.000588509$&$0.000594874$&$1.64\cdot 10^{-7}$\\
SGD 32, $a=0.03$&$0.223260$&$0.000189524$&$0.000533614$&$1.25\cdot 10^{-7}$\\
SGD 32, $a=0.01$&$0.223142$&$2.18864\cdot 10^{-5}$&$0.000588945$&$1.53\cdot 10^{-7}$\\
GD, $a=0.01$&$0.223142$&$2.18084\cdot 10^{-5}$&$0.000552090$&$2.66\cdot 10^{-10}$\\\bottomrule
\end{tabular}
\end{table}

Valleys are selected before the reference center and the subsequent maximum is taken over later saved evaluations. These are retrospective descriptive summaries. The stages overlap in the $a=0.1$ configuration. At $a=0.01$ all 128 model crossings are uncensored. At $a=0.03$ there are 127 and at $a=0.1$ there are 105. The half-fit threshold is $m_*/2$. At $a=0.01$, the finite-amplitude reduction predicts both absolute and relative half-fit times accurately (\cref{tab:twopredictionerrors}). The asymptotic phase-only approximation is saved separately from the finite-amplitude correction. The measured weak gain grows by a factor of $29.92$ during the $a=0.01$ SGD trajectory.

\begin{table}[htbp]\centering\small
\caption{Finite-amplitude prediction errors for the two-layer binary task with 10\% label flips at $a=0.01$. Relative times are centered within each shared stream. MAE for crossing times is measured in slow-time units.}\label{tab:twopredictionerrors}
\begin{tabular}{lrrrr}\toprule
Optimizer & Absolute-time Spearman & Relative-time correlation & Relative-time MAE & Resurgence KL MAE\\\midrule
SGD, batch 32 & $0.99928$ & $0.99681$ & $2.46$ & $3.56\cdot10^{-5}$\\
Full-batch GD & $0.99998$ & $0.9999995$ & $0.0309$ & $1.18\cdot10^{-7}$\\\bottomrule
\end{tabular}
\end{table}

The balanced-count control at $a=0.03$ has endpoint KL $4.36\cdot10^{-6}$. The function-matched single-layer control has endpoint $1.35\cdot10^{-5}$. Both controls remain approximately aligned over the tested post-alignment interval. The batch-one run retains a smaller rise, from $0.000177575$ to $0.000297369$, and endpoint $4.56\cdot10^{-7}$. For $a=0.01$ batch32 SGD, median maximum balance drift over the saved horizon is 19.78, versus 0.970 up to half-fitting. These measurements quantify substantial balance drift alongside accurate timing predictions.

Native PyTorch SGD replays the two dense layers on raw inputs for 4,061 and 3,960 updates at $a=0.03$ and $0.01$. Maximum weight discrepancy is below $7\cdot10^{-16}$, full-network Jacobians reproduce $K$, and 10,000 algebra checks validate the Schur coefficient and discrete balance identity. Pairwise KL is computed directly, with display floors only in log plots. These checks validate the numerical identities and native parameter updates used in the predictions.

\FloatBarrier
\section{Proofs of the phase mechanism and its extensions}\label{sec:twoproofs}
We first prove the phase limit, the discrepancy corollary, and the population-risk identities for deterministic gradient flow and stable full-batch GD. We then extend the GF argument to a predictive weak direction. The finite-step identities below also apply to actual minibatches. The phase proof uses full-batch gradients. We give the phase argument on compact raw-initialization events excluding $u_0+v_{10}=0$. Such events exhaust the Gaussian law. Constants may depend on this event and the realized data.

\subsection{Exact geometry and bounded trajectories}
Write $\mathcal L(s,m)$ for the empirical logistic loss. Its negative derivatives are $A=\bar q-\tfrac 12[\sigma(s+m)+\sigma(s-m)]$ and $B=c-h(s,m)$. Applying the product rule to $s=uv_1$ and $m=auv_2$ gives $\dot{(s,m)}=K(A,B)^T$. Direct multiplication gives the finite-step identities in \cref{eq:twofiniteproducts}. Subtracting the squared factor updates gives \cref{eq:twostochasticbalance}. The terms linear in $\eta$ cancel. The Schur identity follows by putting the terms over $u^2+v_1^2$.

For completeness, with $A$ and $B$ denoting either full or batch residual averages, the raw updates are
\begin{equation}
 u^+=u+\eta(v_1A+av_2B),\quad v_1^+=v_1+\eta uA,\quad v_2^+=v_2+\eta auB.
 \label{eq:twoupdate}
\end{equation}
Their exact effective-coordinate expansion is
\begin{equation}
 \binom{s^+-s}{m^+-m}=\eta K\binom{A}{B}
 +\eta^2\binom{uA(v_1A+av_2B)}{a^2uB(v_1A+av_2B)}.
 \label{eq:twofiniteproducts}
\end{equation}

For gradient flow, $D$ is constant. The effective loss is strictly convex and coercive since both weak groups have clean fractions in $(0,1)$. Its sublevel sets bound $|s|,|m|$. For each fixed $a>0$, \cref{eq:twoU} then bounds every raw parameter. In particular, there is no finite-time blowup. The raw gradient is uniformly continuous on that fixed-$a$ compact set, and its squared norm is integrable by loss descent. Hence the gradient tends to zero.

For GD, start with a bounded effective sublevel set and $|D-D_0|\le 1$. The raw Hessian is bounded by $C/a$ on an enlarged such region: raw parameter norms are $O(a^{-1/2})$, the logit Jacobian contributes $O(1/a)$, and the second logit derivatives are bounded. Choose $\eta\le\nu a$ with sufficiently small $\nu$. The descent lemma gives
\[
 \mathcal L(\theta^+)\le\mathcal L(\theta)-\frac\eta 2\|\nabla_\theta\mathcal L\|^2.
\]
The exact balance correction has absolute value at most $\eta^2\|\nabla_\theta\mathcal L\|^2$, so
\begin{equation}
 \sup_k|D_k-D_0|\le 2\eta[\mathcal L(\theta_0)-\inf\mathcal L]=O(\eta).
 \label{eq:twoDGD}
\end{equation}
Taking $a$ small closes the bootstrap region by induction. This establishes boundedness, descent, and vanishing raw gradients. The descent bound uses the squared norm of the full-batch gradient.

\subsection{Fast strong fitting}
At $a=0$, $v_2$ is fixed and
\[
 \dot u=v_1[\bar q-\sigma(uv_1)],\qquad
 \dot v_1=u[\bar q-\sigma(uv_1)].
\]
The sum and difference have derivatives $\dot{(u+v_1)}=A(u+v_1)$ and $\dot{(u-v_1)}=-A(u-v_1)$. The positive strong optimum is $s_0=\operatorname{logit}(\bar q)$. Except on the stable line $u_0+v_{10}=0$, the strong coefficient converges to $s_0$, with the signs of the limiting factors equal to $\operatorname{sign}(u_0+v_{10})$. The conserved $C_0=u^2-v_1^2$ gives the stated $\widehat u_0$.

On a compact initialization event away from that line, a bounded fast interval brings $s$ into a fixed neighborhood of $s_0>0$. Extending that interval to $O(\log(1/a))$ makes the initial strong error $O(a)$, while the weak weight changes by $O(a\log(1/a))=o(1)$. The slow duration vanishes. The same statement holds for GD because $\eta\le\nu a\to 0$, the strong map is stable, and \cref{eq:twoDGD} makes its balance perturbation vanish.

The strong optimum at a held $m$ is smooth and satisfies $s^*(m)\ge s_0$ and $|s^{*\prime}(m)|<1$. Set $e=s-s^*(m)$. Its exact flow equation is
\begin{equation}
 \dot e=[K_{ss}-s^{*\prime}K_{sm}]A+[K_{sm}-s^{*\prime}K_{mm}]B.
 \label{eq:twotracking}
\end{equation}
On a bounded effective region with $s\ge s_0/2$ and bounded $D$, \cref{eq:twoU} gives $K_{sm}=O(a)$, $K_{mm}=O(a^2+a|m|)$, while $K_{ss}\ge 2s\ge s_0$. Also $A$ has the opposite sign to $e$, with $|A|\ge c_1|e|$ on this compact region. The second term in \cref{eq:twotracking} is $O(a)$ and the first is contracting. Thus $e=O(a)$ after the fast layer, throughout each stopped bounded slow interval. This bound also keeps $s$ positive and removes its lower stopping boundary.

For GD the first-order term is the same. The additional strong-product term is $\eta^2(sA^2+mAB)$, and the Taylor remainder from $s^*(m^+)$ is $O((m^+-m)^2)$. With the stable-step bound, these terms preserve the contracting tracking estimate. The per-update slow duration is $\eta a$. The tracking estimate controls $A$ along the actual training trajectory.

\subsection{Inner dynamics and the seed-dependent amplitude}
Choose the function-preserving sign gauge in which $u$ is positive after the fast layer, and let $w$ be the correspondingly signed weak weight. On bounded $w$ intervals, $m=O(a)$ and strong tracking gives $s=s_0+O(a)$. The balance relation therefore gives $u=\Gamma_{D_0}(w)+o(1)$. In slow time $\tau=a t_{\rm GF}$,
\[
 \frac{\dd w}{\dd\tau}=u[c-h(s,m)]\longrightarrow c\Gamma_{D_0}(w).
\]
This convergence holds on every fixed bounded inner interval by the tracking estimate and ordinary differential-equation stability. In GD the exact increment is $w^+-w=\eta a u B$. \Cref{eq:twoDGD}, vanishing slow mesh and the same tracking estimate give the identical inner limit. Its initial state $w_0$ is fixed by the raw weights.

Writing $\beta(D,w_0)$ for the same integral functional as $\beta(\theta_0)$, the limiting time to a fixed large positive $R$ is $c^{-1}\int_{w_0}^R\dd w/\Gamma_D(w)$. Since
\[
 \Gamma_D(w)=w+\frac{D}{2w}+O(w^{-3})\quad(w\to+\infty),
\]
the limit in \cref{eq:twophasecoef} exists and is finite and positive. This establishes the exponential amplitude $\beta$ of the inner solution.

For matching, continue from $w=R$ until $m=d$, with $R$ large and fixed and $d>0$ small. Strong tracking and bounded balance imply
\[
 \frac{\dd\log w}{\dd\tau}=c+O(w^{-2})+O(|m|).
\]
The logarithmic growth rate is bounded below by $c/2$ after increasing $R$ and decreasing $d$. Therefore the integrated $w^{-2}$ error is $O(R^{-2})$, and the integrated $|m|$ error is $O(d)$. Moreover $m=aw^2[1+O(w^{-2})]$. At the first crossing of $d$,
\begin{equation}
 \tau_d-\tau_a=\frac{\log d-\log\beta(D_0,w_0)}{2c}
 +O(d)+O(R^{-2})+o(1).
 \label{eq:twoexit}
\end{equation}
For GD, summing the Taylor remainders of $\log(w^+/w)$ contributes $O(\eta a\log(1/a))=o(1)$, and crossing overshoots vanish. The $O(\eta)$ balance change affects the inner limit continuously and gives a vanishing error. This yields the same entry phase.

\subsection{Outer dynamics and the three-stage comparison}
On $d\le m\le M_0$, the balance relation gives $K_{mm}/a\to 2m$ uniformly, $K_{sm}=O(a)$, and $A=O(a)$. Consequently
\[
 \frac{\dd m}{\dd\tau}=2mF_0(m)+o(1).
\]
The discrete product remainder divided by the GD slow mesh is
$\eta a[sAB+mB^2]=o(1)$, so the same reduction holds for GD. Gronwall comparison on fixed outer intervals and \cref{eq:twoexit} give \cref{eq:twophase} by taking $a\to 0$, then $R\to\infty$, and then $d\to 0$. Restriction to bounded phase events gives uniformity on any desired shifted-time compact interval.

Implicit differentiation of the conditional strong equation gives
$s^{*\prime}=-(f_+-f_-)/(f_++f_-)$, where $f_\pm=\sigma'(s^*\pm m)>0$. Hence $h(s^*(m),m)$ has derivative $2f_+f_-/(f_++f_-)>0$. Its limit is $\rho$, so $F_0$ has the stated positive zero. The normalized increasing outer orbit is unique, by integrating $1/(2mF_0(m))$ with its logarithmic normalization at zero.

Conditional on $D$, the raw Gaussian construction gives a continuous law for $w_0$. Also
$\partial_{w_0}\log\beta=2/\Gamma_D(w_0)>0$. Thus independent replicas have unequal phases almost surely. At shifted time zero, their limiting weak coefficients differ almost surely. Equation~\eqref{eq:twoKLreduced}, proved below, then gives a strictly positive symmetric KL. At sufficiently early and late fixed shifted times, both coefficients are respectively close to $0$ and $m_*$. Strong tracking makes their logits close too. Initial KL tends to that between independent strong products $u_0v_{10}$ and is positive almost surely. Quantiles of these positive quantities prove \cref{cor:twothree}. Choose the compact initialization event with failure probability below a fraction of $\delta$. Its uniform stable-GD constant $\nu$ then yields the same probability statement after allocating the remaining quantile and limiting errors.

For convergence after the phase interval, early strong fitting already produces empirical loss below $\log 2$ when $a$ is small. Loss descent retains this inequality. Every raw critical point with $u=0$ has effective logits zero and loss $\log 2$, so it cannot be a cluster point. At any critical point with $u\ne 0$, the input-weight gradients force $A=B=0$, hence $(s,m)=(s_*,m_*)$. Boundedness and vanishing gradients therefore imply convergence of the effective logits for both gradient flow and stable GD. Replica KL then vanishes, and the clean-risk formula follows by substituting the two group probabilities.

\subsection{Finite-amplitude reduction and the zero-forcing control}
Following $s=s^*(m)$ in the effective two-dimensional flow requires
$\dot s=s^{*\prime}\dot m$. Eliminating $A$ gives \cref{eq:tworeduction}. This slow-manifold approximation uses raw-initialization inputs after the leading strong fit.

When $c=0$, $B=-h(s,m)$ in gradient flow and
\[
 \frac{\dd v_2^2}{\dd t}=2av_2uB=-2m h(s,m)\le 0.
\]
The weak weight stays bounded. Since $D$ is conserved, $u^2-v_1^2=D+v_2^2$ is bounded. The effective-loss bound on $s=uv_1$ then bounds both factors uniformly in $a$. Thus $|m|=O(a)$ for all time, excluding the nonvanishing weak-phase separation. Finite SGD instead has the second-order term in \cref{eq:twostochasticbalance}. The SGD balanced control numerically tests removal of the weak forcing with this finite-step correction present.

\subsection{Discrepancy, clean risk, and coefficient crossing times}
Let $z_\pm(m)=s^*(m)\pm m$ and $p_\pm(m)=\sigma(z_\pm(m))=\bar q\pm H(m)$. For two coefficients, write $\Delta m=m_0-m_1$, $\Delta s=s^*(m_0)-s^*(m_1)$ and $\Delta H=H(m_0)-H(m_1)$. The exact Bernoulli symmetric-KL identity gives
\begin{align}
 \mathcal S(m_0,m_1)
 &=\frac14\sum_{r\in\{-1,1\}}[\Delta s+r\Delta m]\,r\Delta H\\
 &=\frac12\Delta m\Delta H.
\end{align}
This proves \cref{eq:twoKLreduced}. Its strict positivity for unequal coefficients follows from the positive derivative of $H$. Combined with the phase limit and strong tracking, it also identifies the limiting discrepancy curve explicitly as
\[
 t\longmapsto\mathcal S\big(M(t+\chi_0),M(t+\chi_1)\big).
\]
It approaches zero at both ends of the shifted window, while initial raw strong-product variability gives the separate initial decrease.

On the clean population, sign normalization makes the true-class probability $p_r(m)$. Therefore
\[
 \mathcal R(m)=-\frac12\sum_{r\in\{-1,1\}}\log p_r(m)
 =-\frac12\log[\bar q^2-H(m)^2].
\]
For $m>0$, $H(m)>0$ and $H'(m)>0$, so
\[
 \mathcal R'(m)=\frac{H(m)H'(m)}{\bar q^2-H(m)^2}>0.
\]
At the early and late equilibria, $H(0)=0$ and $H(m_*)=c$. Substitution proves \cref{eq:tworiskgap}. This increase concerns the conditional strong curve, hence the limiting phase trajectory. The late risk itself is the actual GF/GD training limit because the effective logits converge to $(s_*,m_*)$. The classification-error statement follows from $p_+(m_*)=q_+>1/2$ and the sign of $q_--1/2$.

For a fixed threshold $m_{\rm cut}\in(0,m_*)$, the limiting orbit has strictly positive velocity at its crossing. The inner and overlap estimates exclude an earlier macroscopic crossing, and compact-uniform outer convergence brackets the first crossing between the corresponding nearby limiting times. Letting the bracket shrink proves \cref{eq:twohittime}. The first-crossing prediction thus follows from the phase proof and the raw-initialization phase.

\subsection{Phase limit for a predictive weak direction}\label{sec:cleanproof}
This section extends the GF phase argument to a separable, completely uncorrupted task. Its claim concerns compact intervals of shifted time as $a\to0$, with arbitrarily small limiting discrepancy at sufficiently late fixed shifted times. The empirical loss is separable, with the optimum approached as the effective coefficient diverges.

\subsubsection{Clean data, conditional geometry, and residual forcing}
Let $\pi\in(1/2,1)$ be fixed, with a finite design having proportions $\pi/2,\pi/2,(1-\pi)/2,(1-\pi)/2$ on $(b,t)=(+,+),(-,+),(+,-),(-,-)$. When discussing an exact finite design these masses are rational. Set $x=b(1,at)$ and $Y=bt$ for every training and population point. Thus $Y=\operatorname{sign}(x_2)$ for $a>0$. The full linear function class contains a perfect classifier, and no label is flipped or latent. All raw weights $u,v_1,v_2$ are independent unit Gaussians and are trained.

Normalizing logits by $b$ gives $s+tm$ and binary target $T=(1+bY)/2=(1+t)/2$. The loss is \cref{eq:cleanloss}, with negative derivatives
\begin{equation}
 \begin{aligned}
 A_{\rm cl}&=\pi-\pi\sigma(s+m)-(1-\pi)\sigma(s-m),\\
 B_{\rm cl}&=\pi[1-\sigma(s+m)]+(1-\pi)\sigma(s-m).
 \end{aligned}\label{eq:cleangradients}
\end{equation}
The raw GF equations, effective kernel $K$, and conserved balance $D$ in \cref{eq:twokernel,eq:twobalance} are unchanged. For fixed $m$, the strong root is unique. For $m\ge0$ it is
\begin{equation}
 s^*_{\rm cl}(m)=\log\frac{(2\pi-1)e^{-m}
 +\sqrt{(2\pi-1)^2e^{-2m}+4\pi(1-\pi)}}{2(1-\pi)}.
 \label{eq:cleanstrongroot}
\end{equation}
Writing $s_0=\operatorname{logit}(\pi)$, this root decreases from $s_0$ to $s_0/2$. With $f_+=\pi\sigma'(s^*_{\rm cl}+m)>0$ and $f_-=(1-\pi)\sigma'(s^*_{\rm cl}-m)>0$, implicit differentiation gives
\[
 (s^*_{\rm cl})'=-\frac{f_+-f_-}{f_++f_-},\qquad
 J'=-\frac{2f_+f_-}{f_++f_-}<0.
\]
On the conditional curve, $B_{\rm cl}=2J$, $J(0)=\pi(1-\pi)$, and
\begin{equation}
 J(m)\sim\sqrt{\pi(1-\pi)}e^{-m}\quad(m\to\infty).
 \label{eq:cleanfluxtail}
\end{equation}

More generally, for the same reference inputs and binary targets $T_i=(1+b_iY_i)/2$, let $q=\widehat\E T_i\in(1/2,1)$. After fitting the strong coefficient alone, $s_0=\operatorname{logit}(q)$ and
\begin{equation}
 B(s_0,0)=\widehat\E[t_i(T_i-q)]
   =\widehat{\operatorname{Cov}}(t,T).\label{eq:generalforcing}
\end{equation}
In the earlier balanced-group corruption model this empirical residual correlation is $c$. In the deterministic-label task it is the genuine predictive correlation $2\pi(1-\pi)$. Strong tracking, response amplification, and the phase-to-KL comparison together turn this correlation into the three-stage trajectory.

\subsubsection{GF phase limit on compact shifted intervals}
Define $b_0=2\pi(1-\pi)$, $\tau_a=\log(1/a)/(2b_0)$, and let $\Gamma_D$ and $\beta$ be the integral expressions in \cref{eq:twophasecoef} with $s_0=\operatorname{logit}(\pi)$. Put $\chi_j=\log\beta(\theta_{j0})/(2b_0)$.

\begin{proposition}[Clean-label GF phases]\label{prop:cleanphase}
For almost every pair of Gaussian initializations, uniformly on every compact shifted-time interval,
\begin{equation}
 \begin{aligned}
 m_{j,a}((\tau_a+t)/a)&\longrightarrow M_{\rm cl}(t+\chi_j),\\
 s_{j,a}((\tau_a+t)/a)&\longrightarrow
 s^*_{\rm cl}(M_{\rm cl}(t+\chi_j)),
 \end{aligned}\label{eq:cleanphaselimit}
\end{equation}
where $M_{\rm cl}$ is the unique positive orbit in \cref{eq:cleanouter}. It increases from zero to infinity as shifted time ranges from $-\infty$ to $+\infty$.
\end{proposition}
\begin{proof}
Work first on a compact initialization event away from $u_0+v_{10}=0$. The $a=0$ strong equations are exactly those in \cref{sec:twoproofs}, with $\bar q$ replaced by $\pi$. They bring $s$ to $s_0>0$, leave $v_2$ unchanged, and fix the same post-fit sign gauge. A fast interval of order $\log(1/a)$ yields $s=s_0+O(a)$ and weak-weight change $O(a\log(1/a))$.

For the subsequent argument, stop on a fixed bounded effective region containing the conditional curve up to $m=R_0$. The tracking argument proceeds on this stopped region. The root remains at least $s_0/2$ for $m\ge0$ and is positive in a fixed small negative neighborhood of zero. The balance relation bounds the raw factors on the stopped region for each $a$. Smoothness of the weighted sigmoid root gives $|A_{\rm cl}|\ge c_1|s-s^*_{\rm cl}(m)|$ there, with the opposite sign. The exact tracking equation \eqref{eq:twotracking} still holds. Its strong coefficient is contracting and its forcing is $O(a)$, so after the fast layer the tracking error is $O(a)$ throughout the stopped interval. This estimate also prevents exit through the strong-coordinate boundaries.

For bounded signed weak weight $w$, strong tracking yields $s=s_0+O(a)$, $m=O(a)$, and $B_{\rm cl}=b_0+O(a)$. Thus in slow time $\tau=at_{\rm GF}$,
\[
 \frac{\dd w}{\dd\tau}\longrightarrow b_0\Gamma_{D_0}(w).
\]
The inner solution therefore has the same positive amplitude $\beta$. On the overlap from a large fixed $w=R$ to a small fixed $m=d>0$, smoothness gives $B_{\rm cl}=b_0+O(|m|)+O(a)$. The logarithmic growth calculation becomes
\[
 \frac{\dd\log w}{\dd\tau}
 =b_0+O(w^{-2})+O(|m|)+O(a).
\]
As in the earlier proof, the first two integrated errors are $O(R^{-2})$ and $O(d)$. The additional $O(a)$ term integrated over an $O(\log(1/a))$ slow interval vanishes. Hence the crossing phase is
\[
 \tau_d-\tau_a
 =\frac{\log d-\log\beta}{2b_0}
   +O(d)+O(R^{-2})+o(1).
\]
On $d\le m\le R_0$, balance gives $K_{mm}/a\to2m$, $K_{sm}=O(a)$, and tracking gives $A_{\rm cl}=O(a)$. Therefore
\[
 \frac{\dd m}{\dd\tau}=4mJ(m)+o(1)
\]
uniformly on this stopped region. ODE stability and the crossing phase establish the claim on each fixed outer interval. Given a compact shifted-time interval and a bounded phase event, choose $R_0$ beyond its limiting orbit values. Convergence then prevents the weak-coordinate exit. Exhausting compact initialization and phase events proves the almost-sure statement.

The orbit exists uniquely by integrating $1/(4mJ(m))$, whose singularity at zero fixes the exponential normalization. It cannot reach infinity in finite shifted time because \cref{eq:cleanfluxtail} makes this reciprocal nonintegrable at infinity. Its only possible endpoint as shifted time increases is infinity, since its velocity is positive for every finite $m>0$.
\end{proof}

\subsubsection{Replica KL and clean risk}
For points on the conditional curve, $\pi p_++(1-\pi)p_-=\pi$ is independent of $m$. Thus $\pi\Delta p_+=-\Delta J$ and $(1-\pi)\Delta p_-=\Delta J$. Applying the exact Bernoulli symmetric-KL identity gives
\[
 \mathcal S_{\rm cl}
 =\tfrac12[-\Delta J(\Delta s+\Delta m)
          +\Delta J(\Delta s-\Delta m)]
 =-\Delta m\Delta J.
\]
It is strictly positive for unequal coefficients. At the early end both coefficients tend to zero. At the late end, $M_{\rm cl}'(t)=4M_{\rm cl}(t)J(M_{\rm cl}(t))\to0$ by \cref{eq:cleanfluxtail}, so fixed phase shifts give a vanishing coefficient difference and hence vanishing KL. The phase law remains atom-free by the same conditional-density and $\partial_{w_0}\log\beta=2/\Gamma_D(w_0)>0$ argument as before. Independent phases are therefore unequal almost surely.

Consequently, for every $\delta\in(0,1)$ and $\epsilon>0$, there exist deterministic $t_1<0=t_2<t_3$, $g_0>0$, and $a_0>0$ such that for $a<a_0$, with probability at least $1-\delta$, initial KL exceeds its value at $(\tau_a+t_1)/a$, its increase from this time to $(\tau_a+t_2)/a$ is at least $g_0$, and KL at $(\tau_a+t_3)/a$ is below $\min(\epsilon,g_0)$. To see this, first bound initial KL and the positive middle limiting KL below by positive quantiles, then choose early and late shifted times using the two limits above, and finally apply compact-uniform phase convergence. This is a finite-time three-stage statement, with no label corruption.

The true population risk equals \cref{eq:cleanloss} on the matched four-prototype distribution. Along its conditional curve, the envelope derivative is
\[
 \frac{\dd}{\dd m}\mathcal L_{\rm cl}(s^*_{\rm cl}(m),m)
 =-B_{\rm cl}(s^*_{\rm cl}(m),m)=-2J(m)<0.
\]
The risk therefore decreases from the binary entropy of $\pi$ to zero along the limiting orbit, establishing the improving-risk case in Section~\ref{sec:twolayer}.

\clearpage
\section{BatchNorm ablations}\label{sec:bnablation}
We examine normalization and classifier initialization as ablations of the CIFAR-10 ResNet-18 experiment in Section~\ref{sec:setting}. All conditions use 10\% label replacement and the five-epoch warmup recipe in \cref{sec:primarysupp}.

\subsection{Normalization and classifier initialization}\label{sec:configurationablation}
We compare BatchNorm \cite{ioffe2015} and GroupNorm \cite{wu2018} with default or Xavier-uniform classifier weights \cite{glorot2010}, using seeds 123/124 and identical labels and training randomness. Convolution initialization, classifier bias, optimizer, augmentation, and training budget are matched. BatchNorm affine scales start at one and biases at zero. GroupNorm follows \cref{sec:primarysupp}.

\begin{figure}[htbp]\centering
\includegraphics[width=\linewidth]{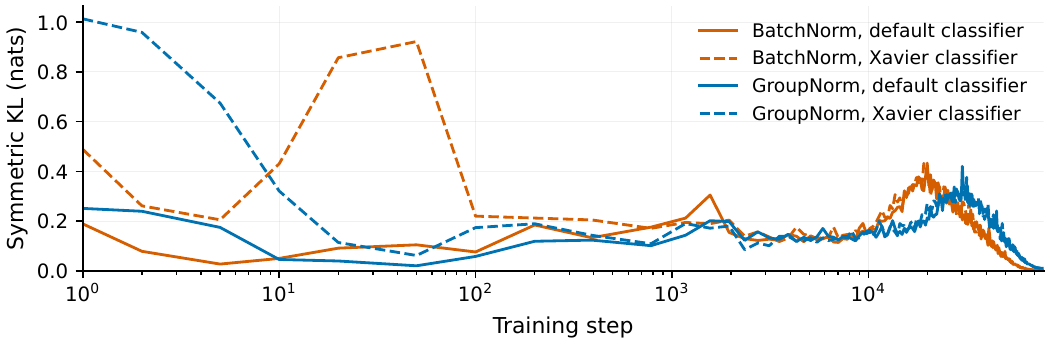}
\caption{Normalization and classifier-initialization ablation for ResNet-18 trained on CIFAR-10 with 10\% label replacement. Curves show unsmoothed full-training-set symmetric KL from step 1 through epoch 200. The four configurations share model seeds 123/124 and the same labels and training recipe.}\label{fig:normalizationablation}
\end{figure}

\Cref{fig:normalizationablation,tab:primaryall} show a resurgence after early agreement in all four configurations. Xavier initialization increases initial discrepancy for both normalization choices. GroupNorm retains the resurgence and produces a later intermediate peak in this comparison.

The exploratory screen used the windows in \cref{tab:primaryall} and required initial/valley at least three, peak/valley at least two, peak-minus-valley at least $0.1$ nat, and final test accuracy within three percentage points of BatchNorm/default. GroupNorm/Xavier was then fixed for the label-ratio comparison and initialization-seed ablation.

\begin{table}[htbp]\centering\small
\caption{CIFAR-10 ResNet-18 with 10\% label replacement. KL is measured in nats at initialization, the valley in epochs 5--30, the subsequent peak through epoch 120, and epoch 200. Accuracy averages the two final test accuracies.}\label{tab:primaryall}
\begin{tabular}{llrrrrr}\toprule
Normalization / classifier&Seeds / replacement&Initial&Valley (epoch)&Peak (epoch)&Final&Accuracy\\\midrule
BatchNorm / default&123/124, 10\%&0.466764&0.114017 (18)&0.433034 (51)&0.001339&89.860\%\\
BatchNorm / Xavier&123/124, 10\%&1.076973&0.117537 (9)&0.431983 (49)&0.001015&90.055\%\\
GroupNorm / default&123/124, 10\%&0.257090&0.117635 (11)&0.420391 (77)&0.009890&89.075\%\\
GroupNorm / Xavier&123/124, 10\%&1.041766&0.084783 (6)&0.376103 (81)&0.009518&89.070\%\\
\bottomrule
\end{tabular}
\end{table}

\FloatBarrier

\subsection{Early running-statistics control}
For BatchNorm ResNet-18 on CIFAR-10 with 10\% label replacement and the default classifier, we compare native training to a branch that keeps its initial weights fixed and updates only running buffers. The two branches share training inputs. The diagnostic uses seed pairs 123/124 and 223/224, batch size 128, the 1,955-step warmup, and a fixed 512-image training probe. Buffer-only weights are verified to remain equal to their initial values.

Over the first five native updates, the buffer-only branch reproduces $99.866\%$ and $99.670\%$ of the corresponding KL drops on this probe. This identifies an early running-statistics effect. The complete trajectories above show the later resurgence under both normalization choices.

\end{document}